\documentclass{article}
\PassOptionsToPackage{numbers, compress}{natbib}
\usepackage[final]{neurips_2022}

\usepackage{amsmath}
\usepackage{amssymb}
\usepackage{mathtools}
\usepackage{amsthm}

\newtheorem{proposition}{Proposition}

\newtheorem{assumption}{Assumption}

\DeclareMathOperator*{\argmin}{arg\,min}
\DeclareMathOperator*{\argmax}{arg\,max}

\usepackage[textsize=tiny]{todonotes}

\usepackage{enumitem}
\usepackage{bm}
\usepackage{algorithm}
\usepackage{algorithmic}

\usepackage[utf8]{inputenc} 
\usepackage[T1]{fontenc}    
\definecolor{mydarkblue}{rgb}{0,0.08,0.45}
\usepackage[colorlinks=true,
            linkcolor=mydarkblue,
            citecolor=mydarkblue,
            filecolor=mydarkblue,
            urlcolor=mydarkblue]{hyperref}
\usepackage{url}            
\usepackage{booktabs}       
\usepackage{amsfonts}       
\usepackage{nicefrac}       
\usepackage{microtype}      
\usepackage{xcolor}         
\usepackage[capitalize,noabbrev]{cleveref}
\usepackage{graphicx}
\usepackage{subfig}

\newcommand*\samethanks[1][\value{footnote}]{\footnotemark[#1]}

\usepackage{titlesec}
\titleformat{\paragraph}
{\normalfont\normalsize\bfseries}{\theparagraph}{1em}{}
\titlespacing*{\paragraph}
{0pt}{0.5ex plus 1ex minus .2ex}{0.5ex plus .2ex}

\title{MissDAG: Causal Discovery in the Presence of Missing Data with Continuous Additive Noise Models}

\author{Erdun Gao\textsuperscript{1}\thanks{Equal contribution.} \ \thanks{Work was done during an internship at JD Explore Academy.} \quad Ignavier Ng\textsuperscript{2}\samethanks[1]  \quad Mingming Gong\textsuperscript{1} \quad Li Shen\textsuperscript{3} \quad \textbf{Wei Huang\textsuperscript{1}} \\ \textbf{Tongliang Liu\textsuperscript{4}} \quad \textbf{Kun Zhang\textsuperscript{2,5}} \quad \textbf{Howard Bondell\textsuperscript{1}}\\
\textsuperscript{1} The University of Melbourne, \textsuperscript{2} Carnegie Mellon University, \\
\textsuperscript{3} JD Explore Academy, \textsuperscript{4} The University of Sydney \\
\textsuperscript{5} Mohamed bin Zayed University of Artificial Intelligence \\
\small{\texttt{erdun.gao@student.unimelb.edu.au}}, \
\small{\texttt{\{ignavierng, kunz1\}@cmu.edu}} \\
\small{\texttt{\{mingming.gong, wei.huang, howard.bondell\}@unimelb.edu.au}} \\
\small{\texttt{shenli100@jd.com}}, \ \small{\texttt{tongliang.liu@sydney.edu.au}} \\
}

\begin{document}

\maketitle

\begin{abstract}
State-of-the-art causal discovery methods usually assume that the observational data is complete. However, the missing data problem is pervasive in many practical scenarios such as clinical trials, economics, and biology. One straightforward way to address the missing data problem is first to impute the data using off-the-shelf imputation methods and then apply existing causal discovery methods. However, such a two-step method may suffer from suboptimality, as the imputation algorithm may introduce bias for modeling the underlying data distribution. In this paper, we develop a general method, which we call MissDAG, to perform causal discovery from data with incomplete observations. Focusing mainly on the assumptions of ignorable missingness and the identifiable additive noise models (ANMs), MissDAG maximizes the expected likelihood of the visible part of observations under the expectation-maximization (EM) framework. In the E-step, in cases where computing the posterior distributions of parameters in closed-form is not feasible, Monte Carlo EM is leveraged to approximate the likelihood. In the M-step, MissDAG leverages the density transformation to model the noise distributions with simpler and specific formulations by virtue of the ANMs and uses a likelihood-based causal discovery algorithm with directed acyclic graph constraint. We demonstrate the flexibility of MissDAG for incorporating various causal discovery algorithms and its efficacy through extensive simulations and real data experiments.
\end{abstract}

\section{Introduction}
Discovering the underlying causal relations among variables of interest often occupies a prominent position for supporting stable inference and rational decisions~\citep{pearl2009causality} in many applications such as medical diagnostics~\citep{richens2020improving}, recommendation systems~\citep{wang2020causal} and economics~\citep{imbens2015causal}. To achieve this goal, conducting randomized controlled trials or using interventions is often acknowledged as the golden rule, which is effective but challenging in practice owing to high costs, ethical issues, or difficulties in obtaining compliance~\citep{resnik2008randomized}. To address this issue, causal discovery from purely observational data, which may be more realistic in specific settings, has drawn considerable attention in both academic and industrial fields~\citep{spirtes2001causation,heckerman2006bayesian, heckerman2008tutorial, glymour2019review}.

Existing causal discovery methods, such as constraint-based methods \citep{spirtes1991pc,colombo2011learning}, score-based methods \citep{chickering2002optimal,raskutti2018learning}, and methods based on functional causal models \citep{shimizu2006linear,hoyer2008nonlinear,zhang2009identifiability,peters2014causal}, typically focus on the settings in which complete observations are available. However, in practice, datasets often suffer from missing values caused by various factors such as entry errors, deliberate non-responses, and sampling drops~\citep{little2019statistical}. Following the definitions by~\citet{little2019statistical}, the missing types can be categorized into three classes, namely missing completely at random (MCAR), missing at random (MAR), and missing not at random (MNAR), according to different missing mechanisms. Many previous efforts have focused specifically on figuring out more identifiable MNAR cases~\citep{bhattacharya2020identification, mohan2021graphical, nabi2020full} and estimating the causal graphs from some specific MNAR cases~\citep{gain2018structure, tu2019causal}, while less attention has been paid to the M(C)AR cases as no extra assumption is required to recover the ground-truth data distribution from incomplete observations~\citep{rubin1976inference, mohan2021graphical}.

\begin{figure}[tb]
\begin{center}
\centerline{\includegraphics[width=0.65\columnwidth]{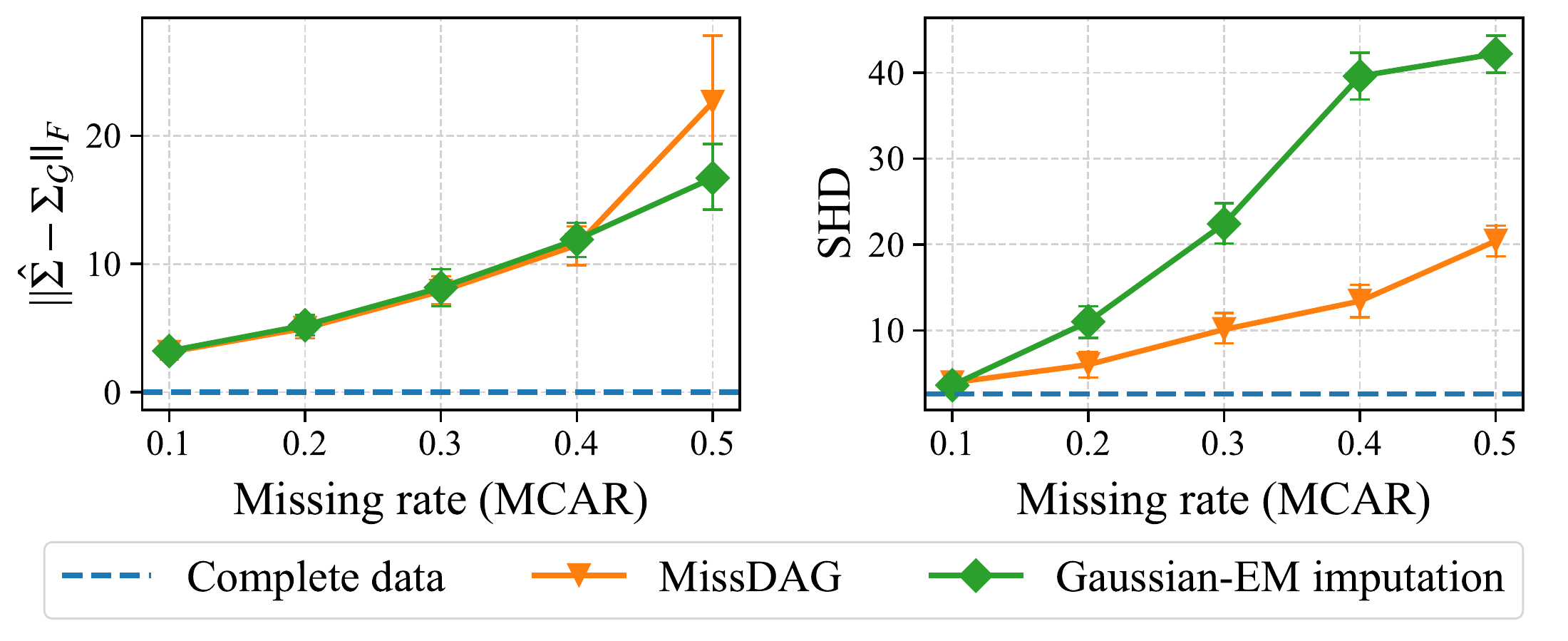}}
\caption{Example of linear Gaussian model with equal noise variance. With zero-mean noises, the recovered covariance matrix $\hat{\Sigma}$ (sufficient statistic) can be a criterion for distribution recovery.}
\label{fig:GEM_VS_MDAG}
\end{center}
\vskip -0.4in
\end{figure}

To perform causal discovery in the M(C)AR case, a naive approach to handle the missing values is the listwise deletion method that simply drops the samples with missing value(s) in at least one of the variables. However, this may lead to unsatisfactory performance if the sample size is limited or the missing rate is high~\citep{stadler2012missing,tu2019causal} because of the decreased statistical power. Another straightforward approach is to impute those missing values instead of listwise deleting them. However, these imputation methods may introduce bias for modeling the underlying data distribution~\citep{kyono2021miracle}. Moreover, as shown in Fig.~(\ref{fig:GEM_VS_MDAG}), even though Gaussian-EM imputation method can consistently recover the data distribution as there is no model misspecification, it may lead to sub-optimal directed acyclic graph (DAG) estimation because it focuses solely on distribution recovery instead of structure learning. Therefore, a principled causal discovery approach that can handle the M(C)AR case is needed. 

\textbf{Contributions.} \ 
In this work, we develop \emph{a practical and general EM-based framework}, called MissDAG, to perform causal discovery in the presence of missing data, in which the underlying missing mechanism is independent from the observed information, which includes the M(C)AR case. Considering the data generating model, we focus on the identifiable additive noise models (ANMs)~\citep{peters2014causal}, including the linear non-Gaussian model~\citep{shimizu2006linear}, linear Gaussian model with equal noise variance~\citep{peters2014identifiability}, and nonlinear ANMs~\citep{hoyer2008nonlinear, peters2014causal}; as a byproduct, our framework also accommodates the typical non-identifiable case, namely the linear Gaussian model with non-equal noise variances that can only be identified to Markov equivalence class~\citep{spirtes1991pc}. The resulting MissDAG framework flexibly accommodates different score based causal discovery algorithms~\citep{chickering2002optimal, zheng2018dags, ng2020role,yuan2011learning} developed for complete data and can be potentially extended to deal with more general cases (e.g., the log-determinant term is incorporated in the likelihood function). Moreover, we conduct extensive experiments on a variety of settings, including synthetic and real data, against many baselines to verify the effectiveness of MissDAG.

\section{Preliminaries}

\textbf{Additive noise models.} \ 
We adopt the notion of structural causal model (SCM)~\citep{pearl2009causality, pearl2016causal} to characterize the causal relations among variables. Each SCM $\mathcal{M}=\langle \mathcal{Z}, \mathcal{X}, \mathcal{F} \rangle$ consists of the exogenous variable set $\mathcal{Z}=\{Z_1, Z_2, \dots, Z_d\}$, the endogenous variable set $\mathcal{X}=\{X_1, X_2, \dots, X_d\}$, and the function set $\mathcal{F}=\{f_1, f_2, \dots, f_d\}$. Here, each function $f_i$ computes the variable $X_i$ from its parents (or causes) $\textbf{Pa}_{X_i}$ and an exogenous variable $Z_i$, i.e., $X_i=f_i(\textbf{Pa}_{X_i}, Z_i)$. In this work, we focus on a specific class of SCMs, called the ANMs ~\citep{hoyer2008nonlinear, peters2014causal}, given by
\begin{equation}
    X_i = f_i(\textbf{Pa}_{X_i}) + Z_i, \quad i=1,2,\dots,d,
    \label{eq:anm}
\end{equation}
where $Z_i$, interpreted as the additive noise variable, is assumed to be independent with variables in $\textbf{Pa}_{X_i}$ and mutually independent with variables in $\mathcal{Z}\backslash Z_i$.

\textbf{Causal graph.} \ 
Each SCM $\mathcal{M}$ induces a causal graph, which we assume in this work to be a DAG\footnote{Indeed, there may be difference between a DAG and a causal graph–the directed edges of the latter is given a causal meaning that allows it to answer interventional queries~\citep{koller2009probabilistic}.}. The DAG $\mathcal{G_{\mathcal{M}}}=(\mathbb{V}, \mathbb{E})$ consists of a vertex set $\mathbb{V}:=\{1,2,\dots,d\}$, in which each node $i$ corresponds to the variable $X_i$, and an edge set $\mathbb{E} \subseteq \mathbb{V}^2$ where $(i,j)\in\mathbb{E}$ if and only if $X_i \in \textbf{Pa}_{X_j}$. Let $X=(X_1,X_2,\dots,X_d)$ be a random vector that includes the variables in $\mathcal{X}$, and $P(X)$ (with density $p(x)$) be the joint distribution of random vector $X$. We assume that there are no latent common causes of the observed variables (i.e., \emph{causal sufficiency}), which, together with the \emph{acyclicity} assumption, indicates that $P(X)$ and induced DAG $\mathcal{G_{\mathcal{M}}}$ satisfy the causal Markov condition \citep{pearl2009causality}.
\section{Problem definition}

\textbf{Notations.} \ 
In this paper, we focus on the finite-sample setting with missing data. Consider a dataset $\mathcal{D} = (\mathbf{X}, \mathbf{Y})$ that consists of $N$ samples, where $\mathbf{X}\in \mathbb{R}^{N\times d}$ and $\mathbf{Y}\in \mathbb{R}^{N\times d}$. Each row $(\mathbf{X}_i, \mathbf{Y}_i)$, independently sampled from $P(X,Y)$, represents the $i$-th observation. $\mathbf{Y}$ is the indicator matrix that records the missing positions in $\mathbf{X}$, i.e., $\mathbf{Y}_{ij}=0$ if $\mathbf{X}_{ij}$ is missing and $\mathbf{Y}_{ij}=1$ otherwise. In the presence of missing data, the fully observed data $\mathbf{X}$ is unavailable. For simplicity, let $\mathbf{O}$ group all observed positions in $\mathbf{X}$ and $\mathbf{M}$ group all the missing positions. For each observation $\mathbf{X}_i$, let $\mathbf{o}$ group the indexes of the observed part and $\mathbf{m}$ group the indexes of the missing part of $\mathbf{X}_i$. Notice that $\mathbf{o}$ and $\mathbf{m}$ are different for different observations. Then, $\mathbf{X}_{\mathbf{O}}=[\mathbf{X}_{ij}:\mathbf{Y}_{ij}=1, i\in [N], j\in [d]]$ includes all observed positions and $\mathbf{X}_{\mathbf{M}}=[\mathbf{X}_{ij}:\mathbf{Y}_{ij}=0, i\in [N], j\in [d]]$\footnote{We use $[N] = \{1,2,\dots,N\}$ to represent the set of all integers from $1$ to $N$.} includes all missing positions in $\mathbf{X}$. Similarly, we have $\mathbf{X}_{i\mathbf{o}}=[\mathbf{X}_{ij}:\mathbf{Y}_{ij}=1, j\in [d]]$ and $\mathbf{X}_{i\mathbf{m}}=[\mathbf{X}_{ij}:\mathbf{Y}_{ij}=0, j\in [d]]$.


Following~\citep{little2019statistical}, we define the full likelihood of the $i$-th sample $(\mathbf{X}_{i\mathbf{o}}, \mathbf{Y}_i)$ as
\begin{equation*}
\label{eq:full_likelihood}
\begin{aligned}
\mathcal{L}_{\rm full}(\mathbf{X}_{i\mathbf{o}}, \mathbf{Y}_i; \theta, \psi) =& \int p(\mathbf{X}_{i\mathbf{o}},\mathbf{X}_{i\mathbf{m}}; \theta) \ p(\mathbf{Y}_i |\mathbf{X}_{i\mathbf{o}}; \psi)\ \mathrm{d}\mathbf{X}_{i\mathbf{m}}, \\
 =& p(\mathbf{Y}_i |\mathbf{X}_{i\mathbf{o}}; \psi) \int p(\mathbf{X}_{i\mathbf{o}},\mathbf{X}_{i\mathbf{m}}; \theta)\ \mathrm{d}\mathbf{X}_{i\mathbf{m}},
\end{aligned}
\end{equation*}
where the parameters $\psi$ govern the missing mechanisms and $\theta$ include all the model parameters. The ignorable likelihood of $\mathbf{X}_{i\mathbf{o}}$ is defined as
\begin{equation*}
\label{eq:ignorable_likelihood}
    \mathcal{L}_{\rm ignorable}(\mathbf{X}_{i\mathbf{o}}; \theta) = \int p(\mathbf{X}_{i\mathbf{o}}, \mathbf{X}_{i\mathbf{m}}; \theta)\ \mathrm{d}\mathbf{X}_{i\mathbf{m}}.
\end{equation*}
\begin{assumption}[Ignorable missingness \citep{little2019statistical}]
\label{ass:Ignorable_Missingness}
The inference about parameters $\theta$ based on the ignorable likelihood evaluated only by $\mathbf{X}_{i\mathbf{o}}$ is the same as inference for $\theta$ based on the full likelihood.
\end{assumption}

Ignorable missingness is important as it a common assumption that is required by EM-style algorithms and also our method. It can be interpreted as a belief that the available data is sufficient to "correct" the missing data, by assuming that the missingness and model parameters are distinct.

\textbf{Task definition.} \ 
Consider a distribution $P(X, Y)$, where the marginal distribution $P(X)$ is induced from a SCM $\mathcal{M}$ satisfying the assumption of ANM as defined in Eq.~(\ref{eq:anm}), and a dataset $\mathcal{D}=(\mathbf{X}, \mathbf{Y})$. In practice, the observed distribution is only $P(X_o, Y)$ satisfying Assumption~\ref{ass:Ignorable_Missingness} and the observational part of the dataset is $\widetilde{\mathcal{D}} = (\mathbf{X}_{\mathbf{O}}, \mathbf{Y})$. Our task is to learn the DAG $\mathcal{G_{\mathcal{M}}}$ from the dataset $\widetilde{\mathcal{D}}$.
\section{MissDAG}
In this section, we introduce our proposed method, called MissDAG, which leverages the penalized EM framework to iteratively identify the causal graph $\mathcal{G_{\mathcal{M}}}$ and model parameters from the incomplete data $\widetilde{\mathcal{D}}$. In the M-step, MissDAG takes the log-likelihood of the observational part of the sample and applies a penalty function as the score function to guide the search of model parameters. Instead of directly modeling the complex likelihood of the sample, MissDAG equivalently models the simpler noise distributions by the density transformation~\citep{mooij2011causal} since ANMs always limit the noise distributions to some specific distribution families. Moreover, the prior information of causal structure is also considered as an inductive bias to reduce the variance of parameter estimation. In the E-step, the log-likelihood function is integrated over the posterior of missing entries to obtain the expectation if the closed-form of the posterior is available. Otherwise, Monte Carlo (MC) simulations~\citep{wei1990monte} are adopted to numerically compute the expectation. The details are shown in the following subsections.

\subsection{The overall framework of MissDAG}
Leveraging the development of score based causal discovery methods, we follow the style that taking the log-likelihood of observations (only the observational part) and a penalty function as the score function. Then, the general form of the optimization problem can be written as
\begin{equation*}
\label{eq:overall_score}
    \begin{aligned}
    \arg \max_{\theta} \ \ & \mathcal{S}(\theta) = \log p(\mathbf{X}_{\mathbf{O}}; \theta) - \lambda\ \textbf{PEN}(\theta), \\
    \text{subject to} \ \ &  \mathcal{\theta}_{\mathcal{G}} \in \textbf{DAGs},
\end{aligned}
\end{equation*}
where $\theta=(\theta_{\mathcal{G}}, \theta_{\mathcal{F}})$, including a graph learning part $\theta_{\mathcal{G}}$ and a causal mechanisms learning part $\theta_{\mathcal{F}}$, denotes the parameters of an SCM $\mathcal{M}$. However, in some methods~\citep{zheng2018dags,zheng2020learning,ng2020role}, $\theta_{\mathcal{G}}$ can be absorbed into and induced from $\theta_{\mathcal{F}}$. \textbf{PEN}$(\cdot)$ is the penalty function and $\lambda$ is the penalty coefficient. With the i.i.d. assumption of each observation $(\mathbf{X}_i, \mathbf{Y}_i)$, the score function $\mathcal{S}(\theta)$ can be written as 
\begin{equation}
    \mathcal{S}(\theta)= \sum_{i=1}^N \log \int p(\mathbf{X}_{i\mathbf{o}}, \mathbf{X}_{i\mathbf{m}}; \theta)\ \mathrm{d}\mathbf{X}_{i\mathbf{m}} - \lambda\ \textbf{PEN}(\theta).
\end{equation}
Unfortunately, the closed-form solution of $\mathcal{S}(\theta)$ cannot be obtained. Since the score function in Eq.~(\ref{eq:overall_score}) gives rise to a penalized maximum log-likelihood estimation problem, we can take the iterative penalized EM method~\citep{chen2014penalized}, which relates the parameters estimation of the SCM from $\log p(\mathbf{X}_{\mathbf{O}}; \theta)$ to the same parameters estimation from the complete-data log-likelihood $\log p(\mathbf{X}; \theta)$.

Different from the imputation methods that replace the missing entries $\mathbf{X}_{\mathbf{M}}$ by some specific values, EM based methods formulate a two-step iterative operation. We start with an initial value $\theta^0$ and denote $\theta^t$ as the estimate of $\theta$ at the $t$-th iteration. Then, each iteration of the EM method can be represented as the following two steps:
\begin{itemize}[leftmargin=*]
\item \textbf{E-step} takes the estimated model parameters of the previous step and the observational part $\mathbf{X}_{\mathbf{O}}$ to impute the missing entries by the distribution of $\mathbf{X}_{\mathbf{M}}$, which, in other words, gets the expected log-likelihood of the complete-data as follows.
\begin{equation}
\begin{aligned}
    \mathcal{Q}(\theta, \theta^t) =&\int p(\mathbf{X}_{\mathbf{M}} | \mathbf{X}_{\mathbf{O}}; \theta^t) \log p(\mathbf{X}_{\mathbf{O}}, \mathbf{X}_{\mathbf{M}}; \theta)  \ \mathrm{d}\mathbf{X}_{\mathbf{M}} - \lambda\ \textbf{PEN}(\theta) \\
                                  =& \mathbb{E}_{X_m|X_o; \theta^t} \left\{ \log p(\mathbf{X}_{\mathbf{O}}, \mathbf{X}_{\mathbf{M}}; \theta) \right\} - \lambda\ \textbf{PEN}(\theta).
\end{aligned}
\label{eq:Q-function}
\end{equation}

\item \textbf{M-step} calculates $\theta^{t+1}$ by maximizing the $\mathcal{Q}$ function as follows.
\begin{equation*}
    \theta^{t+1} \in \arg\max_\theta \mathcal{Q}(\theta, \theta^t)
\end{equation*}

\end{itemize}

The E-step calculates the expected data log-likelihood $\mathcal{Q}(\theta, \theta^t)$. The following M-step, then, maximizes $\mathcal{Q}(\theta, \theta^t)$ in $\theta$ for the fixed $\theta^t$ with DAG constraint. The convergence analysis of MissDAG is provided in Appendix~\ref{app:convergence_analysis}. 

\subsubsection{Log-likelihood term of $\mathcal{S}(\theta)$}
The final problem comes to the exact formulation of $p(\mathbf{X}; \theta)$, which is not straightforward to obtain in our problem since ANMs typically impose assumptions on the noise distribution instead of the joint distribution $P(X)$. Benefiting from the well-researched results of the change of variables rule of density transformation~\citep{mooij2011causal}, we can equivalently formulate $p(\mathbf{X}; \theta)$ by transforming it to $p(\mathbf{Z}; \theta)$. Here, for simplicity, we take $Z=(Z_1,Z_2,\dots,Z_d)$ and $f=(f_1,f_2,\dots,f_d)$. Then, we have
\begin{equation}
\label{eq:pd_transformation}
p_X(X) = p_Z(X-f(X)) |\det (\mathbf{I} - \mathbf{J}_f)|,
\end{equation}
where $\mathbf{I}$ is the identity matrix and $\mathbf{J}_f$ represents the Jacobian of $f$ evaluated on $X$. 

\begin{proposition}
\label{lemma:det_equals_1}
 If $\theta_{\mathcal{G}}$ represents a DAG, then $|\det (\mathbf{I} - \mathbf{J}_f)| = 1$.
\end{proposition}
The proof is included in Appendix~\ref{app:proofs_det_equals_1}. With Proposition~\ref{lemma:det_equals_1}, the log-determinant term becomes zero if the candidate solution is acyclic and can be dropped to simplify Eq.~(\ref{eq:likelihood_transformation}). However, we would also like to point out that some recent score-based structure learning methods based on continuous optimization, e.g., GOLEM~\citep{ng2020role} and NOTEARS-ICA~\citep{zheng2020thsis}, have shown that including the log-determinant term (which corresponds to likelihood based on directed cyclic graphs) in its objective function may be desirable and lead to better performance. In this work, this term is ignored since only acyclic models are considered. Then, with Eq.~(\ref{eq:pd_transformation}) and Proposition~\ref{lemma:det_equals_1}, we can model the log-likelihood of the simpler mutually independent noises distributions. Then, we have
\begin{equation}
\label{eq:likelihood_transformation}
    \mathcal{L}(\mathbf{X}_i; \theta) = \sum_{j=1}^{d} \log p_{Z_j}(\mathbf{X}_{ij}-f_j(\mathbf{X}_i)) +\log |\det (\mathbf{I} - \mathbf{J}_f)| = \sum_{j=1}^{d} \log p_{Z_j}(\mathbf{X}_{ij}-f_j(\mathbf{X}_i)).
\end{equation}

In the E-step, with different data generation models, the exact formulation of the posterior may be unavailable. In the next section, therefore, we split up these two cases named exact posterior and approximate posterior respectively for presentations.


\section{Different posterior cases}

\subsection{Exact posterior}
Firstly, we deal with the linear Gaussian models including linear Gaussian model with equal variance (LGM-EV) and linear Gaussian model with non-equal variance (LGM-NV). That is to say, in Eq.~(\ref{eq:anm}), each $f_i \in \mathcal{F}$ is a linear function and $Z_i \sim \mathcal{N}(0, \sigma_i^2)$ with $\Sigma_Z = \text{diag} (\sigma_1^2, \sigma_2^2, \dots, \sigma_d^2)$. Then, the model can be rewritten as 
\begin{equation}
\label{eq:linear_anm}
    X = W^T X + Z,
\end{equation}
where $W\in \mathbb{R}^{d\times d}$ is the weight matrix and $W_{ij} \neq 0$ means that $X_i$ is one of the causes of $X_j$. Based on the density of a linear transformation, we know $P(X)$ belongs to a multivariate Gaussian distribution. For multivariate Gaussian, sufficient statistics consist of the mean vector $\mu(X)$ (the first-order moment) and the covariance matrix $\operatorname{cov}(X)$ (the second-order moment). With zero-mean assumption of $Z$, we have $\mu(X) = \mathbf{0}$ and estimate $\operatorname{cov}(X)$ by $\mathbf{T}$. In other words, we can equivalently replace $p(\mathbf{X}; \theta)$ by $p(\mathbf{T}; \theta)$. Specifically, with full data, $\mathbf{T}$ can be directly calculated by $\mathbf{T} =\frac{1}{N} \mathbf{X}^{T} \mathbf{X}$. 

In this case, we specify $\theta=(W, \Sigma_Z)$ since the two parameters can govern the distribution $P(X)$. According to Eq.~(\ref{eq:likelihood_transformation}), the complete log-likelihood $\mathcal{L}(\mathbf{X}; W, \Sigma_Z)$ can be sufficiently formulated by
\begin{align}
\label{eq:linear_sufficient_formulation}
\mathcal{L}(\mathbf{X}; W, \Sigma_Z) &= \log p(\mathbf{T}; W, \Sigma_Z) \notag \\  &= -\frac{1}{2}\text{Tr}(\log \Sigma_Z) -\frac{1}{2N} \text{Tr}((\mathbf{I}-W)^T\mathbf{T}(\mathbf{I}-W)\Sigma_Z) -\frac{d}{2} \log 2\pi.
\end{align}

Since (\ref{eq:linear_sufficient_formulation}) is linear in $\mathbf{T}$, the $\mathcal{Q}$ function can be formulated in a closed-form.

\subsubsection{\textbf{E-step:} compute $\mathcal{Q}(W, \Sigma_Z, W^t, \Sigma_Z^t)$} 
As discussed before, the E-step calculates the expected log-likelihood $\log p(\mathbf{T}; W, \Sigma_Z)$ with $\mathbf{X}_{\mathbf{O}}$ and $(W^t, \Sigma_Z^t)$. From Eq.~(\ref{eq:linear_anm}), we have $X=(\mathbf{I}-W^T)^{-1}Z$. Then, the implicit parameter $\Sigma_X^t$ can be estimated by $\Sigma_X^t=(\mathbf{I}-W)^{-T}\Sigma_Z^t(\mathbf{I}-W)^{-1}$. Then, $\mathbf{T}^t=\mathbb{E}[\mathbf{T} \ | \ \mathbf{X}_{\mathbf{O}}; \Sigma_X^t]$ can be straightforwardly calculated by the well-known results on the conditional distributions of the multivariate Gaussian. Each entry of $\mathbf{T}^t$ is obtained by $\mathbf{T}^t_{ij} = [\frac{1}{N}\sum_{k=1}^N \xi_k \ | \ \mathbf{X}_{\mathbf{O}}; \Sigma_X^t]$, where 
\begin{equation*}
    \xi_k
    = \begin{cases}
    {\Sigma_X^t}_{ij} + \hat{\mathbf{X}}_{ki} \hat{\mathbf{X}}_{kj} & \text{if } \mathbf{X}_{ki} \& \mathbf{X}_{kj} \text{ are missing}, \\
    \hat{\mathbf{X}}_{ki} \hat{\mathbf{X}}_{kj} & \text{otherwise}.
    \end{cases}
\end{equation*}
$\hat{\mathbf{X}}_k$ records the expectation of missing part of the $k$-th instance $\mathbf{X}_k$ and also shares the same indexes with $\mathbf{X}_k$. $\hat{\mathbf{X}}_k$ is initialized as $\mathbf{X}_k$. Then, $\hat{\mathbf{X}}_{k\mathbf{m}} = {\Sigma_X^t}_{\mathbf{m}\mathbf{o}} {\Sigma_X^t}_{\mathbf{o}^{-1}\mathbf{o}} \mathbf{X}_{k\mathbf{o}}$.

\subsubsection{\textbf{M-step:} maximize $\mathcal{Q}(W, \Sigma_Z, W^{t}, \Sigma_Z^{t})$}

With $\mathbf{T}^t$, the M-step maximizes the score $\mathcal{S}(\theta)$ with $\mathcal{L}(\mathbf{T}^t; W, \Sigma_Z) = \log p(\mathbf{T}^t; W, \Sigma_Z)$. Here, we also plug in the DAG constraint for optimization to ensure that the estimated graph is acyclic. Then, the overall optimization problem can be formulated as
\begin{align*}
    W^{t+1}, \Sigma_Z^{t+1} = & \ \argmax_{W, \Sigma_Z}  \ \  \mathcal{L}(\mathbf{T}^t; W, \Sigma_Z) +  \lambda\ \textbf{PEN}(W), \\
     \text{subject to} & \ \mathcal{G}_W \in \textbf{DAGs}, \nonumber
\end{align*}
where $\mathcal{G}_W$ means the graph induced from $W$. Here, we do not restrict the use of any specific algorithm for solving this problem. For the LGM-EV, one can adopt the greedy search method by \citet{peters2014identifiability}, GOLEM~\citep{ng2020role}, or NOTEARS~\citep{zheng2018dags} to estimate the DAG $\mathcal{G}$, while for the LGM-NV, one can apply different search methods like GES~\citep{chickering2002optimal}, A*~\citep{yuan2011learning}, and GOLEM~\citep{ng2020role}.

\subsection{Approximate posterior}
Unfortunately, for Non-Linear (NL)-ANMs and Linear Non-GAussian Model (LiNGAM) cases, the E-step is not available in a closed-form, which would make the likelihood inference for missing data more difficult. There are mainly two problems to compute the expectation: (1) computing the posterior distribution $p(\mathbf{X}_{i\mathbf{m}} | \mathbf{X}_{i\mathbf{o}}; \theta^t)$; (2) computing the integral. Since the problem mainly comes from the noise modeling as we later show, we prefer to first introduce the M-step of our method.

\subsubsection{\textbf{M-step:} maximize $\mathcal{Q}(\theta, \theta^t)$}
From  Eq.~(\ref{eq:likelihood_transformation}), we know that two reasons lead to the non-closed-form for integral. The first one is that $f$ includes some complex non-linear functions. However, considering the likelihood issue, the non-linearity problem can be well handled by taking neural networks to model $f$~\citep{zheng2020learning, ng2019masked, lachapelle2020gradient}. Therefore we leave $f$ for both linear and non-linear models for brevity. The second problem is non-Gaussian noise, which needs to be clarified before the M-step.

\textbf{The problem of $p_Z(Z)$.}  \ \ \
If noises are non-Gaussian, the exact formulation of each noise distribution is unknown. To make the likelihood-based methods work, a fixed Super (Sub)-Gaussian prior distribution can be set to model the noise distributions~\citep{hyvarinen2010estimation, zheng2020learning}. The theoretical result that the maximum likelihood estimate is locally consistent even in the presence of small misspecification error is well-established~\citep{amari1997stability}. Here, we take the Super-Gaussian distribution as an example:
\begin{equation}
\label{eq:super-G}
p_i(Z_i) = c_Z\exp{(-2\log \cosh{(Z_i)})},
\end{equation}
where $c_Z$ is a constant. Notice that using (\ref{eq:super-G}) needs to assume the unit scale of noise, which may not hold for the real scenarios. Therefore, we prefer to take a normalized likelihood for standardized noise variable $Z_i/\sigma_i$. Then, the standardized log-likelihood\footnote{The term ``standardized log-likelihood'' is taken to follow the literature of independent component analysis.} of the observation $\mathbf{X}_i$ will be 
\begin{align}
\label{eq:standardized_likelihood}
    \mathcal{L}_{\rm standardized}(\mathbf{X}_i; \theta) =& \sum\limits_{j=1}^d \log p_{Z_j}\left(\frac{\mathbf{X}_{ij} - f_j(\mathbf{X}_i)}{\hat{\sigma}_j}\right) - \log \hat{\sigma}_j,
\end{align}

where $\sigma_j^2=\text{Var}(Z_j)$ is the variance of noise $Z_j$ with empirical version of $\hat{\sigma}_j^2= \frac{1}{N}\sum_{j=1}^N (\mathbf{X}_{:,j} - f_j(\mathbf{X}))^2$. Then, Eq.~(\ref{eq:standardized_likelihood}) can serve as the log-likelihood term and the overall optimization problem in the M-step can be written as 
\begin{align*}
    \theta^{t+1} = & \ \argmax_{\theta} \ \  \mathcal{L}_{\rm standardized}(\mathbf{X}^t; \theta) - \lambda\ \textbf{PEN}(\theta), \\
    \text{subject to} & \  \theta_{\mathcal{G}} \in \textbf{DAGs}, \nonumber
\end{align*}
where $\mathcal{L}_{\rm standardized}(\mathbf{X}^t; \theta)$ represents the expected log-likelihood function in the $t$-th iteration.

\subsubsection{\textbf{E-step:} compute $\mathcal{Q}(\theta, \theta^t)$}
\textbf{The problem of the integral.}  \ 
For LiNGAM and NL-ANMs, there is no closed-form for the integral to get the expectation of log-likelihood in the E-step. Naturally, we take the Monte Carlo sampling method~\citep{robert2004monte} to approximate the expectation. Then, Eq.~(\ref{eq:Q-function}) can be reformulated as 
\begin{equation*}
\begin{aligned}
\mathcal{Q}(\theta, \theta^t) = \mathbb{E}_{X_m|X_o; \theta^t} \left\{ \log p(\mathbf{X}_{\mathbf{O}}, \mathbf{X}_{\mathbf{M}}; \theta) \right\} = \sum_{i=1}^N \sum_{j=1}^{N_s} \log p(\mathbf{X}_{i\mathbf{o}}, x^j_{i\mathbf{m}}; \theta),
\end{aligned}
\end{equation*}
where $x^j_{i\mathbf{m}}$, sampled from the posterior, represents the $j$-th value of the total $N_s$ sampling results for the missing part $\mathbf{X}_{i\mathbf{m}}$ of the observation $\mathbf{X}_i$. 

\textbf{The problem of sampling from the posterior.}  \
Instead of directly sampling from $p(\mathbf{X}_{i\mathbf{m}} | \mathbf{X}_{i\mathbf{o}}; \theta^t)$ to fill the missing part $\mathbf{X}_{i\mathbf{m}}$, we use rejection sampling~\citep{bishop2006PRML} to sample from a proposal distribution $Q(X_{\mathbf{m}})$ with the probability density function $q(X_{\mathbf{m}})$, from which we can readily draw samples. Then, a constant $c_k$ is set to guarantee that $c_k q(X_{\mathbf{m}}) \geq p(\mathbf{X}_{i\mathbf{m}} | \mathbf{X}_{i\mathbf{o}}; \theta^t)$ for $\forall X_{\mathbf{m}}$. For each sample $x_{\mathbf{m}}^j$ from $q(X_{\mathbf{m}})$, the accept rate would be $p_{\rm accept}=p(x_{\mathbf{m}}^j| \mathbf{X}_{i\mathbf{o}}; \theta^t) / c_k q(x_{\mathbf{m}}^j)$. However, $p(\mathbf{X}_{i\mathbf{m}} | \mathbf{X}_{i\mathbf{o}}; \theta^t)$ can not be directly obtained. With Bayes Rule, we have
\begin{equation*}
    p(\mathbf{X}_{i\mathbf{m}} | \mathbf{X}_{i\mathbf{o}}; \theta^t) = \frac{p(\mathbf{X}_i; \theta^t)}{p(\mathbf{X}_{i\mathbf{o}}; \theta^t)},
\end{equation*}
while for each instance $\mathbf{X}_i$, $p(\mathbf{X}_{i\mathbf{o}}; \theta^t)$ is a constant marked as $c_o$. However, the joint distribution $p(\mathbf{X}_i; \theta^t)$ is still not directly available. With Eq.~(\ref{eq:pd_transformation}) and Proposition \ref{lemma:det_equals_1}, we find that we may skip for obtaining the closed-form of $p(\mathbf{X}_i; \theta^t)$ but equivalently provide the value of $p_Z(\mathbf{Z}_i; \theta^t)$:
\begin{equation*}
    p(\mathbf{X}_i; \theta^t) = |\det (\mathbf{I} - \mathbf{J}_{\theta^t_f})| p_Z(\mathbf{Z}_i; \theta^t) = \prod_{j=1}^{d} p_{Z_j}(\mathbf{X}_{ij}-{\theta^t_{f_j}}(\mathbf{X}_i)).
\end{equation*} 
Then, just like in the M-step, we take a normalized distribution to model the noise distributions. Then, the probability that a sample can be accepted is given by
\begin{equation*}
    p_{\rm accept} = \frac{\prod_{j=1}^{d} p_{Z_j}(\mathbf{X}_{ij}-{\theta^t_{f_j}}(\mathbf{X}_i))}{c_o c_k q(x_{\mathbf{m}}^j)}.
\end{equation*}
Here, we take $c_r=c_o c_k$ as a re-normalized constant to activate the rejection sampling methods.

\section{Experiments}

We report the empirical results to verify the effectiveness of MissDAG on both synthetic and a biological dataset.

\textbf{Baselines.} \ 
We mainly take imputation methods as baselines including Mean Imputation, MissForest Imputation~\citep{stekhoven2012missforest}, and Optimal Transport (OT)-imputation~\citep{muzellec2020missing} to impute the incomplete data at first,\footnote{Results on more imputation methods including GAIN~\citep{yoon2018gain}, KNNImputer~\citep{andridge2010review}, MICE~\citep{white2011multiple} are shown in Appendix~\ref{app:more_baselines}.} and then apply the causal discovery methods including GOLEM~\citep{ng2020role}, NOTEARS~\citep{zheng2018dags}, the algorithm (`Ghoshal') by \citet{ghoshal2018learning}, NOTEARS-MLP~\citep{zheng2020learning}, and NOTEARS-ICA~\citep{zheng2020thsis} to estimate the causal graph for different assumed models (see Appendix~\ref{app:solving_optimization} for details.). For LGM, we also include the structural EM method for multivariate Gaussian distribution, which we called Gaussian-EM imputation, to recover the complete data. For LGM-NV that aims to identify the Complete Partial DAG (CPDAG), we also include different searching methods such as A*~\citep{yuan2011learning} and GES~\citep{chickering2002optimal} to solve the optimization problem of the M-step. Also, we include the Test-wise Deletion PC (TD-PC) as a baseline for LGM-NV. For LiNGAM, we also use ICA-LiNGAM and Direct-LiNGAM as baseline methods and put the results in Appendix~\ref{app:different_cd_methods}. The detailed implementations of the imputation and structure learning methods, as well as the hyper-parameters of the proposed method, are presented in Appendix~\ref{app:implementation_details}.

\textbf{Metrics.} \ 
We report the widely used criterion named Structural Hamming Distance (SHD), which refers to the smallest number of edge additions, deletions, and reversals required to transform the recovered DAG into the true one, averaged over $10$ random repetitions to assess how the edges differ between the estimated and ground-truth DAG in the identifiable cases. For the non-identifiable case such as LGM-NV, we report the SHD-CPDAG to measure the distances between different CPDAGs. Other criteria such as F$1$ and recall are included for the supplementary experiments in Appendix~\ref{app:supplementary_experiments}.

\begin{figure*}[ht]
\centering
\subfloat[LGM-EV.]{\label{fig:different_miss_rates_LGM_EV}
  \includegraphics[width=0.465\textwidth]{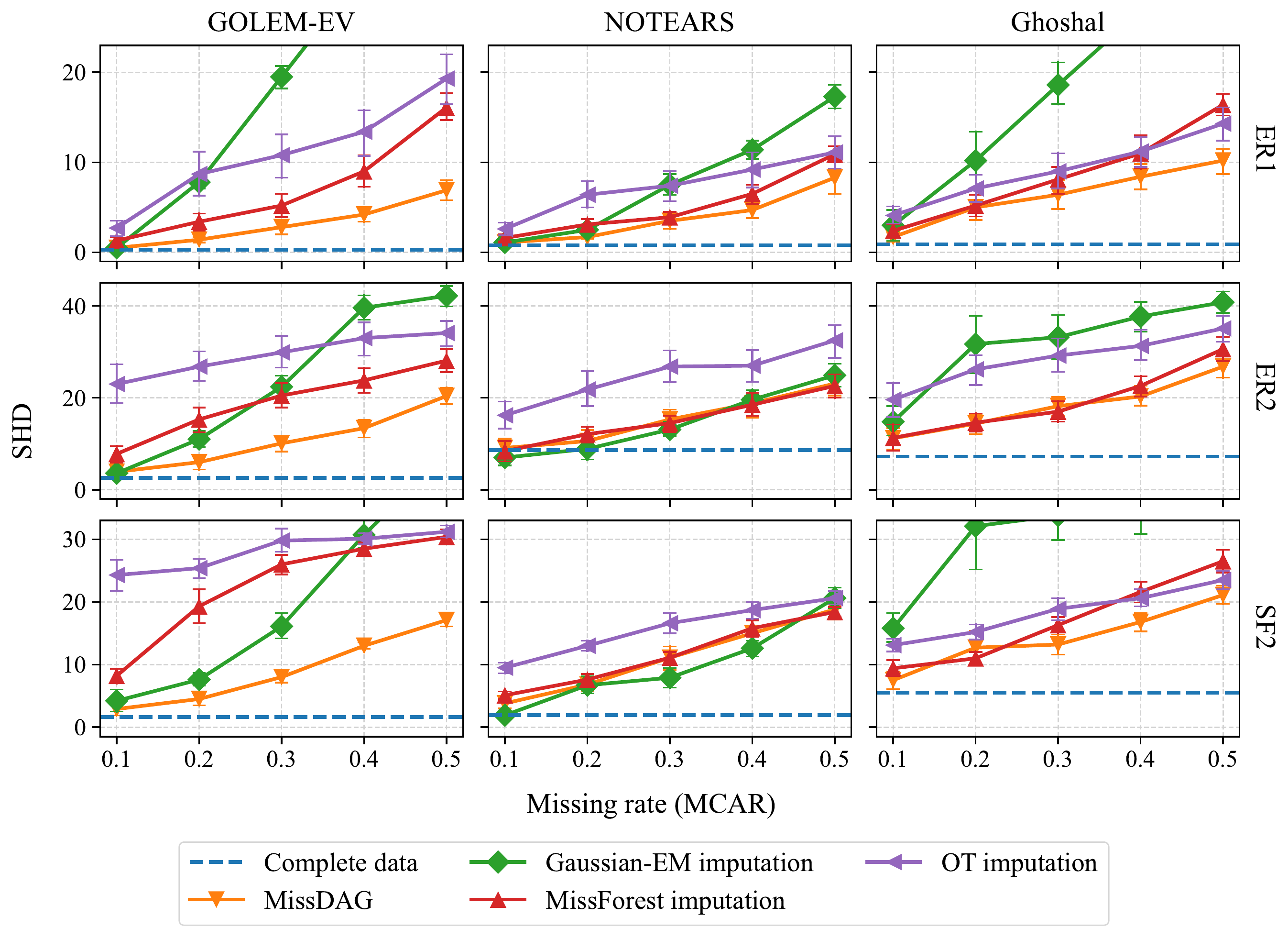}
}
\subfloat[LGM-NV.]{\label{fig:different_miss_rates_LGM_NV}
  \includegraphics[width=0.465\textwidth]{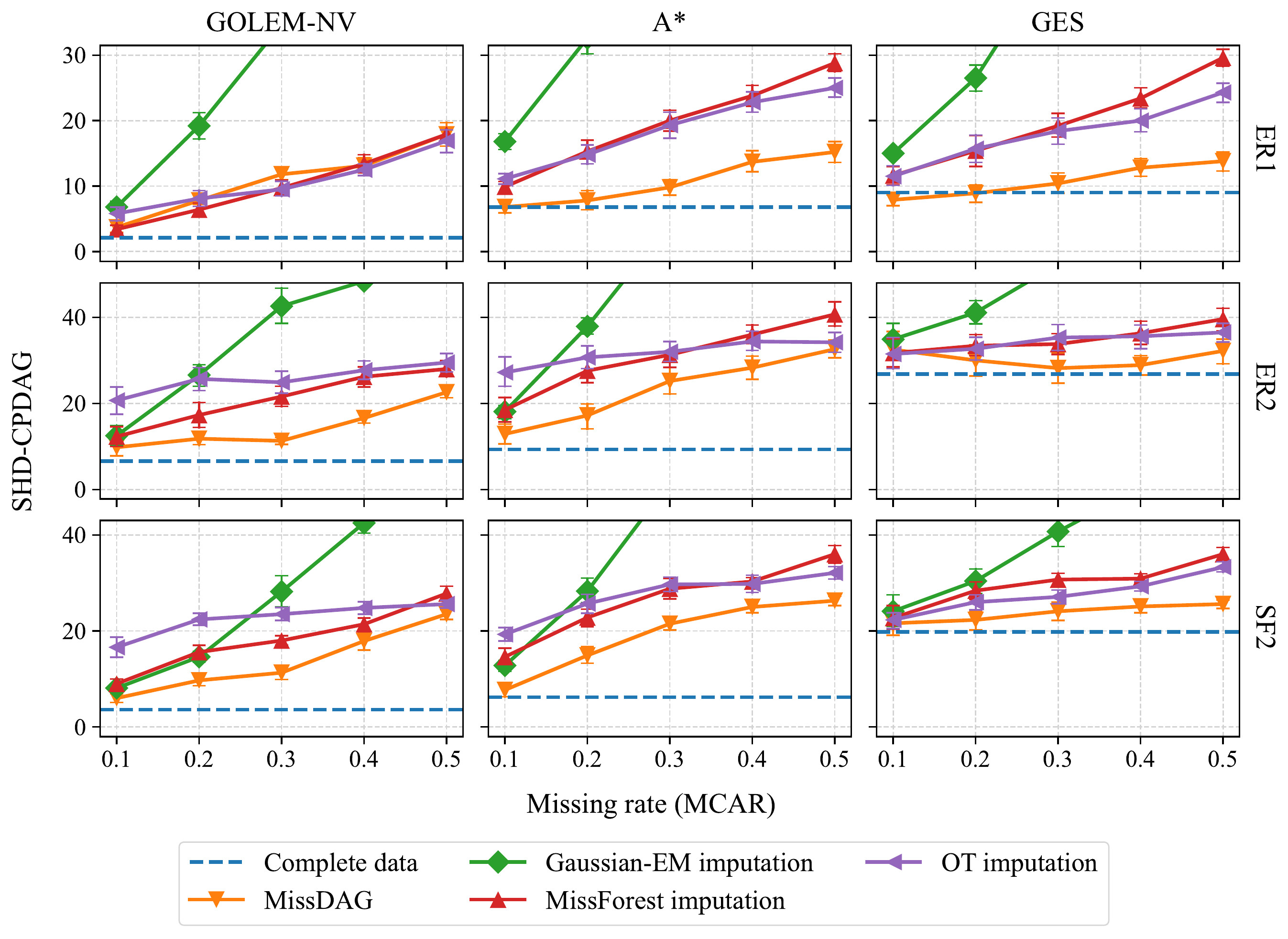}
}
\caption{Recovery of the true structure measured by SHD or SHD-CPDAG ($\downarrow$). (a) LGM-EV with $d=20$. (b) LGM-NV with $d=15$ since the searching time of A* is too long. Rows: ER1, ER2, and SF2 graphs. Columns: different methods. Some results for Gaussian-EM imputation are truncated because its SHDs are too large in those cases.}
\label{fig:different_miss_rates_LGM}
\vskip -0.1in
\end{figure*}

\textbf{Simulations.} \ 
The synthetic data we consider here is generated according to the ANM in Eq.~(\ref{eq:anm}). As illustrated above, we consider four cases, including LGM-EV, LGM-NV, LiNGAM, and NL-ANM. In each experiment, a ground-truth DAG $\mathcal{G}$ with $d$ nodes and $kd$ directed edges was first generated from one of the two graph models, Erd{\H{o}}s-R{\'e}nyi (ER) or Scale-Free (SF). According to different edges, $kd$ edges $(k=1,2)$, the graph model is named ER$k$ or SF$k$. We also simulate denser graphs in Appendix~\ref{app:different_degrees}. Then, for linear models, a weighted matrix $W \in \mathbb{R}^{d\times d}$ with coefficients sampled from $\operatorname{Uniform}([-2.0,-0.5]\cup[0.5,2.0])$ with equal probability is generated to assign values to each edge in $\mathcal{G}$. For the non-linear model, corresponding to each edge in $\mathcal{G}$, a $f_i$ is constructed from a fixed MLP with random coefficients. The non-Gaussian noise we take here follows a Gumbel distribution. In line with the settings outlined in~\citep{zheng2020thsis}, we do not consider the scenario of non-equal scales, as the normalized likelihood makes the optimization problem hard to be solved. Our framework, however, can be extended with future advanced methods to tackle this challenge. Experimental results on more different non-linear functions and different noise distributions are included in Appendix~\ref{app:different_non_linear_types} and Appendix~\ref{app:different_noise_types}, respectively. In equal variance/scale sets, all independent noises belongs to their distributions with variance/scale as $1$ while non-equal variance/scale settings get the scale of each noise independently sampled from $\operatorname{Uniform} [1.0,2.0]$. For each experiment, we sample $100$ observations for linear models and $200$ observations for non-linear models. More results on different numbers of samples with fixed number of nodes and different numbers of nodes with the fixed number of samples are shown in Appendix~\ref{app:different_numbers_of_samples} and Appendix~\ref{app:different_numbers_of_nodes}, respectively. We also add the experiment with $d$ nodes and $2d$ observations with run-time comparisons in Appendix~\ref{app:scalability_of_different_nodes}. The missing type in our experiments is MCAR while the results on MAR and MNAR are also provided in Appendix~\ref{app:different_missing_types}.

\textbf{Linear Gaussian case.}  \
In Fig.~(\ref{fig:different_miss_rates_LGM_EV}), across all settings for LGM-EV, including different graphs and missing rates, MissDAG with GOLEM, NOTEARS and Ghoshal as baseline methods can show consistently the best performance or performance comparable to the best performances. While all imputation methods are sensitive to different baseline methods, all NOTEARS-based methods show improvements compared to GOLEM-based methods although GOLEM is the real full likelihood method for LGM. However, 
GOLEM solves the problem by soft-constraint, which may suffer from the finite sample. For imputation methods, we can see that MissForest usually acquires the second-best place. Gaussian-EM imputation can consistently recover the multivariate Gaussian distribution, but its performance varies a lot with different causal discovery methods. With NOTEARS, Gaussian-EM imputation can gain better, even the best results as compared to the others. The results of LGM-NV are shown in Fig.~(\ref{fig:different_miss_rates_LGM_NV}). Three different searching strategies are considered. It is observed that MissDAG can still achieve the best performances across all settings. The capacities of A* and GES may be limited by the finite observations while GES appears to perform the worst. Moreover, A*, an exact search method, may suffer from high computing complexity. The comparisons on running time are provided in Appendix~\ref{app:running_time}.

\begin{figure*}[h]
\centering
\subfloat[Ground truth.]{
  \includegraphics[height=0.16\textwidth]{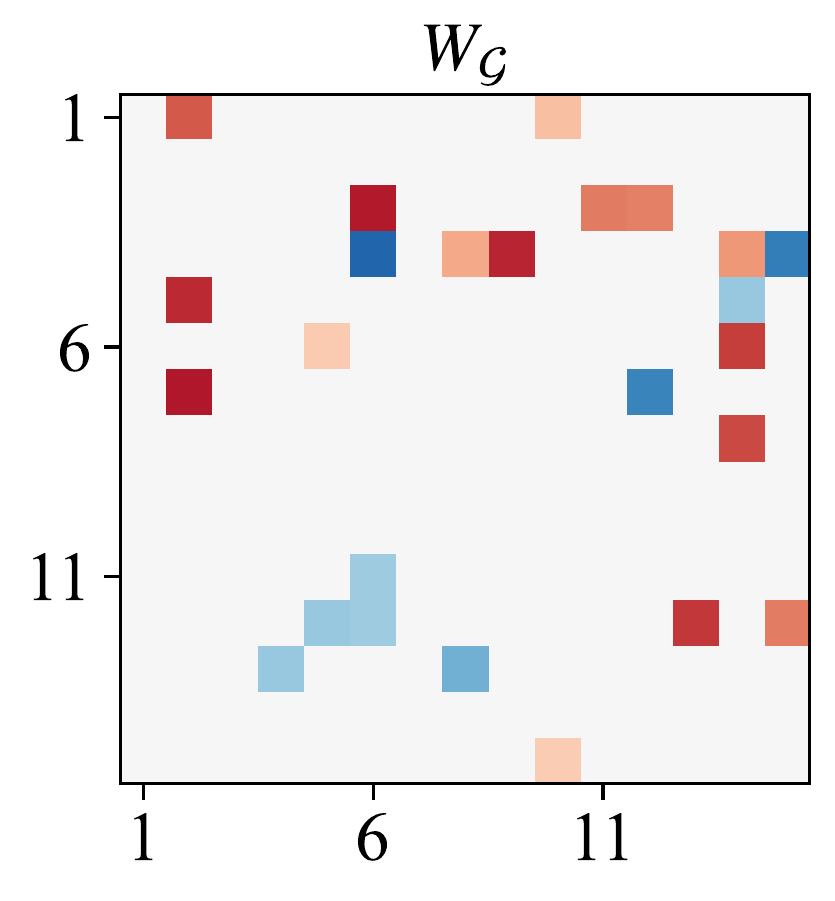}
}
\subfloat[Optimization process; $W^{(n)}$ refers to estimated graph in $n$-th EM iteration.]{
  \includegraphics[height=0.16\textwidth]{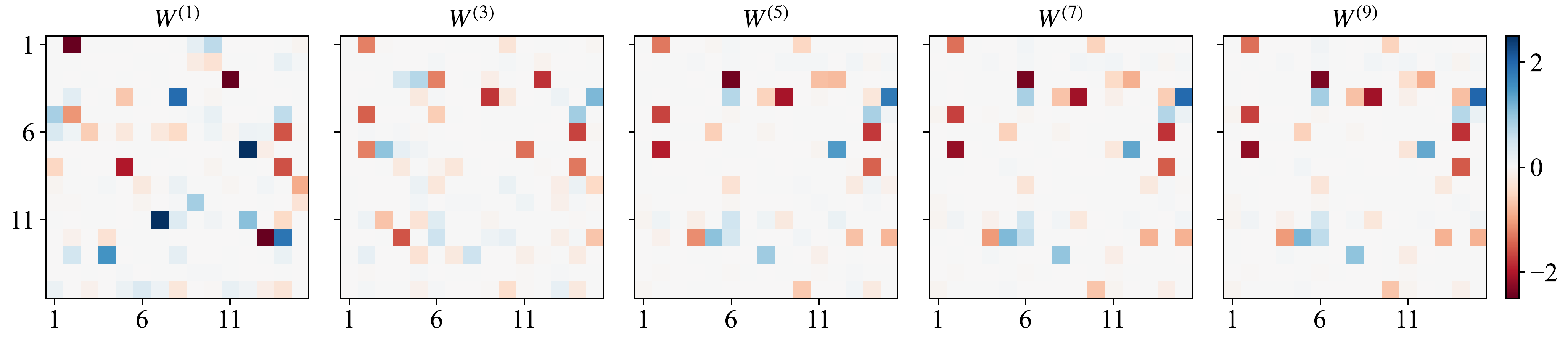}
} \\
\subfloat[LGM-EV.]{
  \includegraphics[width=0.85\textwidth]{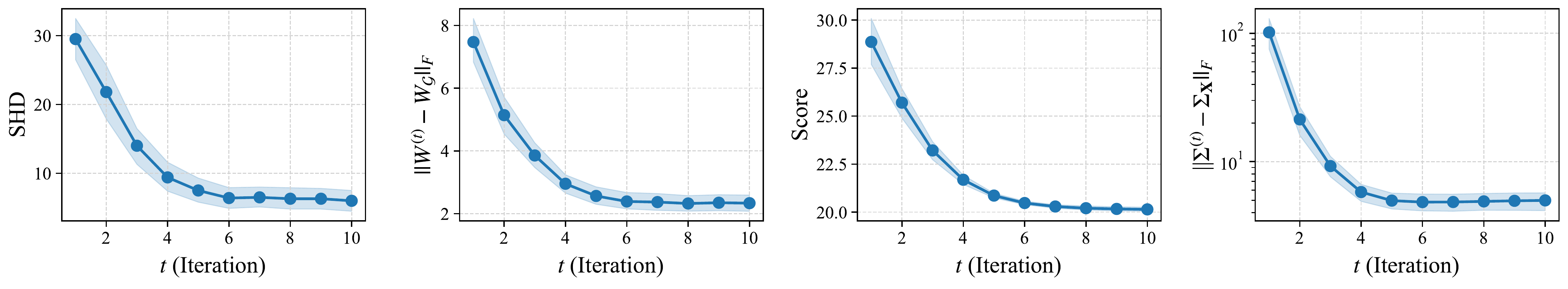}
}
\caption{Visualizations of the optimization process. }
\label{fig:missdag_optz_process}
\vskip -0.12in
\end{figure*}

\textbf{Visualization of the learned DAG of MissDAG.}  \ 
We take an example of the MissDAG optimization process on LGM-EV and plot the change in estimated parameters in Fig.~(\ref{fig:missdag_optz_process}), which shows that the learned causal graph asymptotically approximates the ground-truth DAG $\mathcal{G}$, including the existence of edges and their weights. The data distribution can also be well recovered.


\begin{figure*}[h]
    \begin{center}
    
    	
    	\minipage{0.63\textwidth}
        	\subfloat[LiNGAM.]{\label{fig:different_miss_rates_non_gaussian}
            \includegraphics[width=\linewidth]{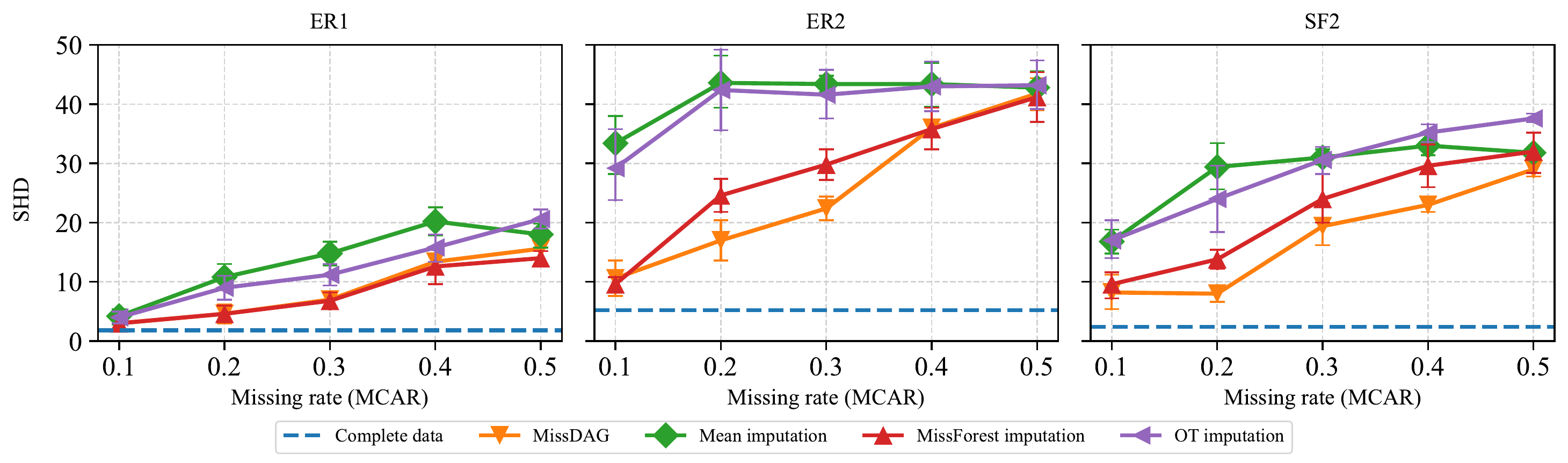}}\\
        \vskip -0.01in
            \subfloat[Non Linear ANMs.]{\label{fig:different_miss_rates_non_linear}
            \includegraphics[width=\linewidth]{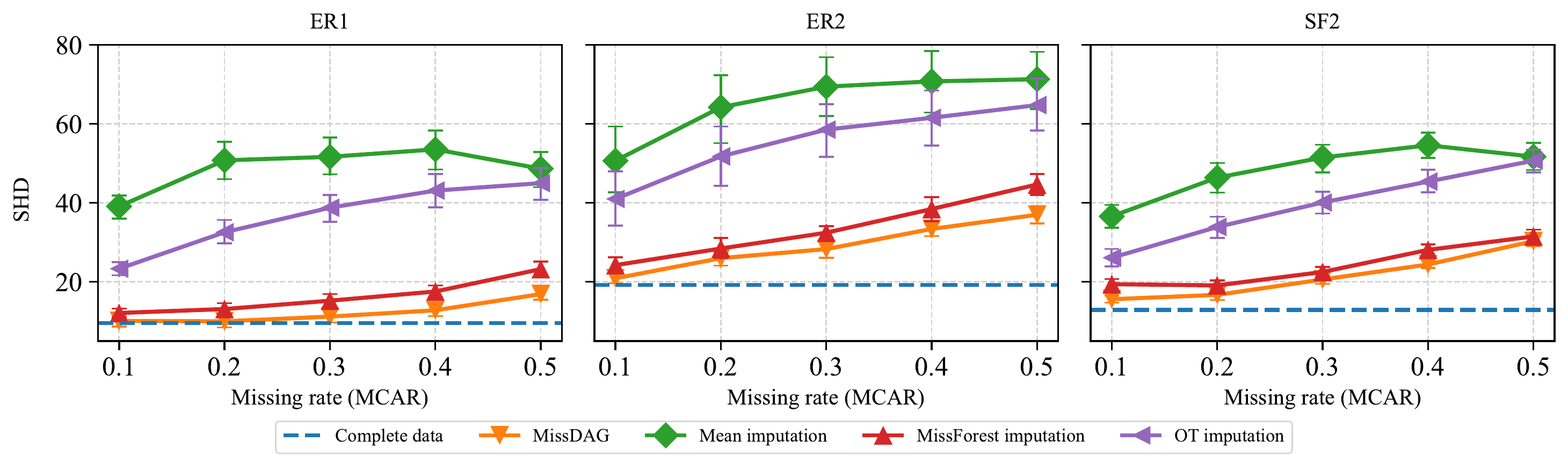}}
    	\caption{MissDAG with approximate posterior.}
    	\endminipage
    	\minipage{0.32\textwidth}
        	\subfloat[LGM.]{\label{fig:real_data_lgm_shd}
            \includegraphics[width=\linewidth]{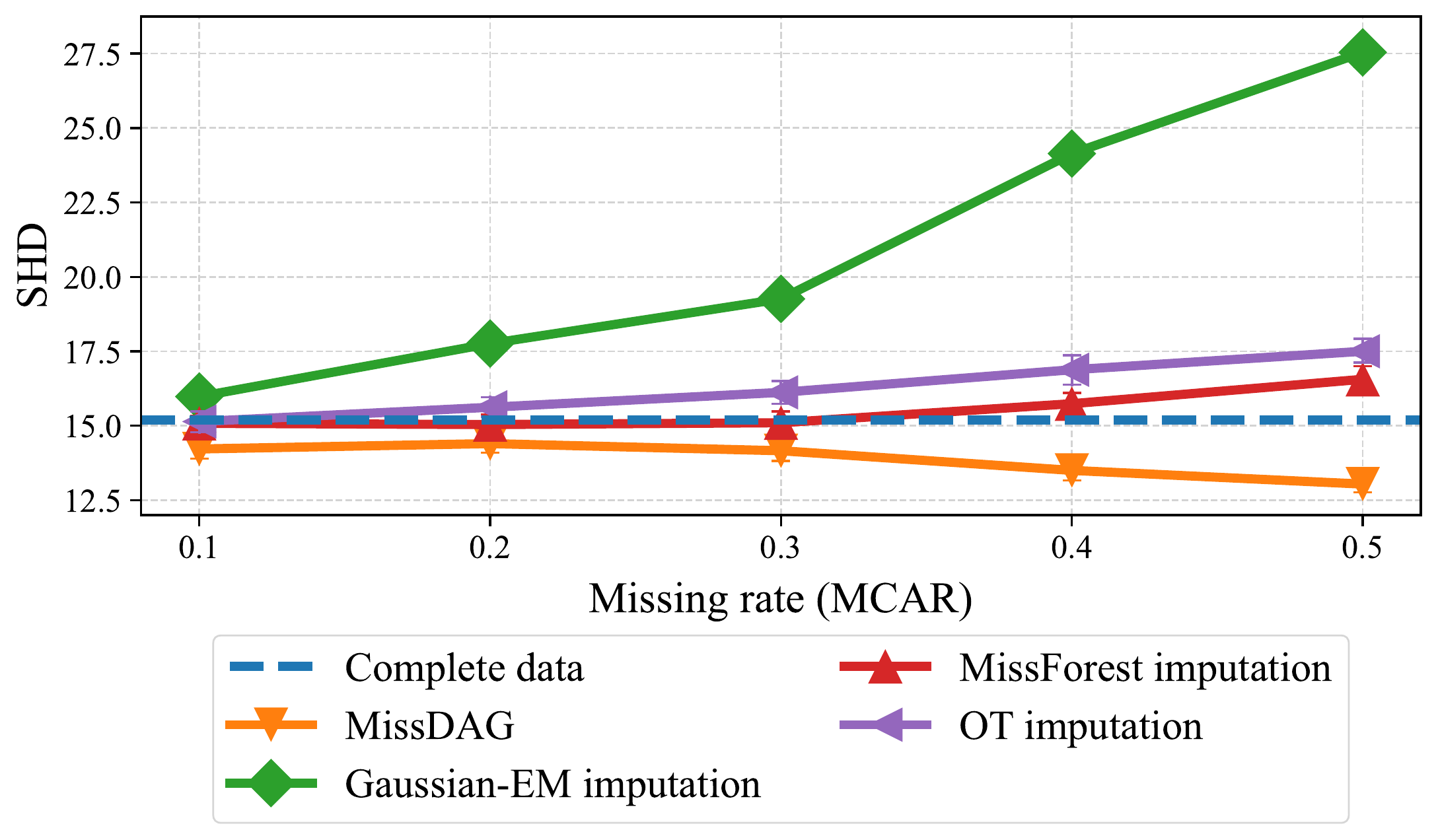}
            }\\
        \vskip -0.01in
            \subfloat[LiNGAM.]{\label{fig:real_data_lingam_shd}
            \includegraphics[width=\linewidth]{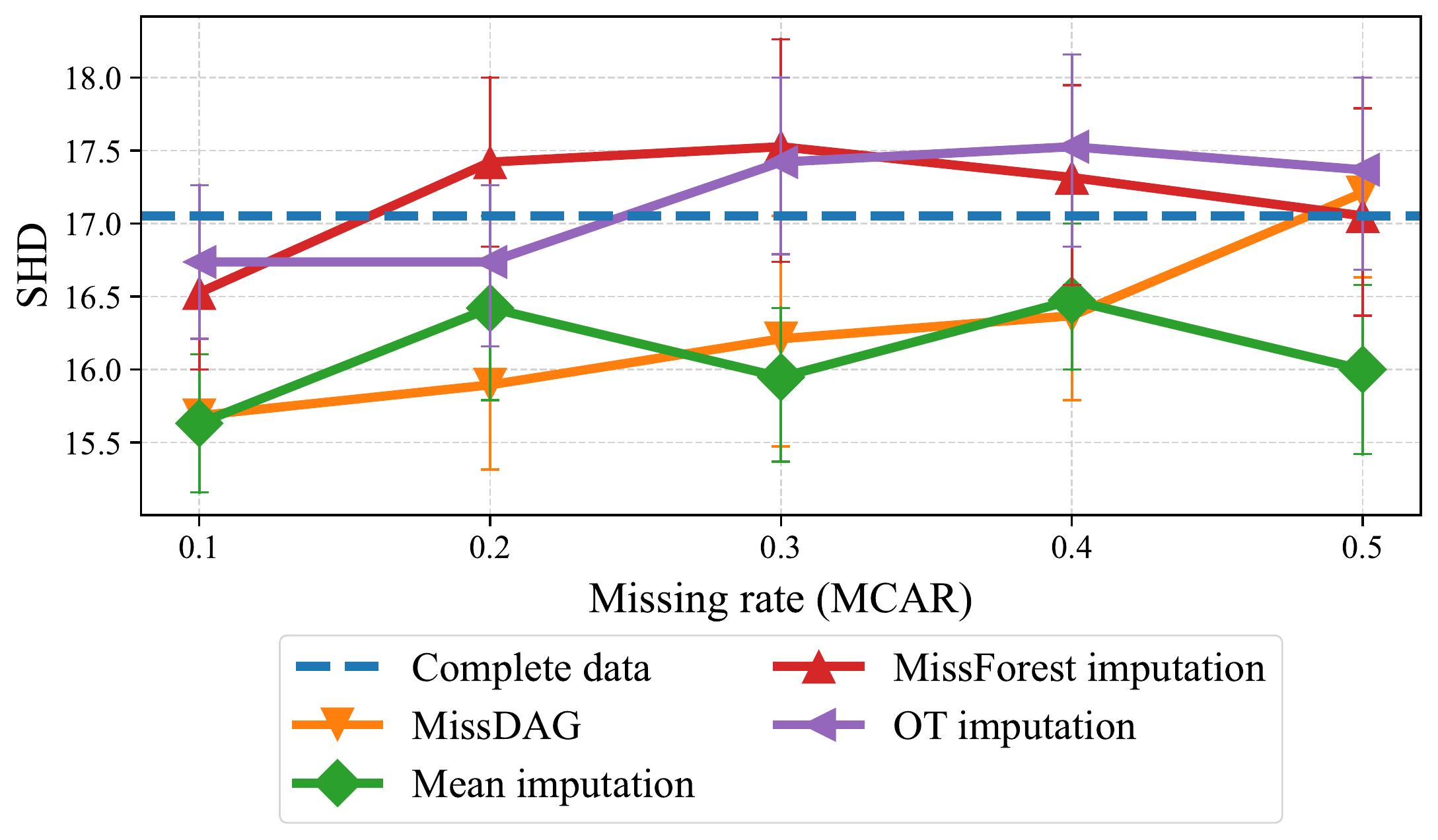}}
    	\caption{Dream4 results.}\label{fig:real_data_shd}
    	\endminipage
	
	\end{center}
	\vspace{-1.4em}
\end{figure*}

\textbf{LiNGAM and NL-ANMs.}  \ 
Fig.~(\ref{fig:different_miss_rates_non_gaussian}) shows the performances of MissDAG and different baseline methods with NOTEARS-ICA as the causal discovery algorithm on LiNGAM while Fig.~(\ref{fig:different_miss_rates_non_linear}) provides the results on NL-ANM of $20$ variables. One can see that MissDAG always occupies the best or one of the best methods across all settings. MissForest~\citep{stekhoven2012missforest} always shows the best or the second-best performance. Especially for NL-ANM, MissForest can get comparable results to MissDAG. 

\textbf{Biological dataset.}  \ 
We take a biological dataset named Dream4 and provided in~\citep{greenfield2010dream4}, which simulates gene expression measurements from five sub-networks of transcriptional regulatory networks of E. coli and S. cerevisiae. Here we consider the 10-node networks, which however include feedback loops. We can see that MissDAG occupies the first place for most of the settings and LGM is more suitable to learn from this dataset. The results in Fig.~(\ref{fig:real_data_shd}) show that there are misspecifications between our models and real data, probably due to the cycles in the real data.

\section{Related works}

\textbf{Causal discovery from complete data.} \ 
Two lines of methods prevail in causal discovery research, namely constraint-based methods, such as PC and fast causal inference (FCI)~\citep{spirtes2001causation}, and score based methods like GES~\citep{chickering2002optimal}. The first branch reads the (conditional) independencies information encoded in the data distribution, which can also be viewed as an equality constraint~\citep{shpitser2014introduction}, to decide the existence of edges and directions of some edges. However, the constraint-based method can only reach the Markov Equivalence Class (MEC) of the ground-truth DAG since DAGs in the same MEC share the totally same conditional independencies information. The second branch searches the model parameters in the DAG space by maximizing the penalized likelihood (score) on the observational data. For a long time, these methods suffer from the high searching complexities of combinatorial optimization. Recently, NOTEARS~\citep{zheng2018dags} recasts this problem as a continuous optimization by introducing an algebraic characterization of DAG. Then, NOTEARS has been extended to handle non-linear cases ~\citep{yu2019dag, ng2019masked, zhu2020causal, lachapelle2020gradient, wang2021ordering}, time-series data~\citep{pamfil2020dynotears}, unmeasured confounder~\citep{bhattacharya2021differentiable} and interventional data~\citep{brouillard2020differentiable}. By imposing further assumption on the data generating model, NL-ANMs~\citep{hoyer2008nonlinear, peters2014causal}, Post-Nonlinear Model (PNL)~\citep{zhang2009identifiability}, LGM-EV~\citep{peters2014identifiability}, and LiNGAM~\citep{shimizu2006linear}, etc, are proposed to learn the ground-truth DAG with identifiability guarantees. 

\textbf{Causality with incomplete data.} \ 
MissDeepCausal~\citep{mayer2020missdeepcausal} leverages the deep latent model to estimate the causal effects of a treatment, intervention or policy from incomplete data. GINA~\citep{ma2021identifiable} systematically analyzes the identifiability of generative models under MNAR case and designs a practical deep generative model which can provide identifiability guarantees for certain MNAR mechanisms. DECI~\citep{geffner2022deep} proposes a general deep latent model to perform both causal discovery and inference. Moreover, the theoretical results can guarantee that this model can identify the causal graph under standard causal discovery assumptions.

\textbf{BN learning with incomplete data.} \ 
Previous methods~\citep{heckerman2008tutorial, singh1997learning, friedman1997learning, friedman1998bayesian, riggelsen2006learning} mainly inherit the Expectation-Maximization (EM) method framework \citep{rubin1976inference}, which conducts likelihood inference in an iterative optimization way. \citet{friedman1997learning, singh1997learning} iteratively refine the conditional distributions and sampling the missing values from these distributions. \citep{tanner1987calculation} proposed a data augmentation method by a stochastic simulation-based method that draws the filled-in value from a predictive distribution. The augmentation method was accelerated by recasting the problem into two phases, parent set identification by an exact search and structure optimization by an approximate algorithm~\citep{adel2017learning}. However, existing BN learning methods from incomplete data focus on identifying the Markov equivalence classes (i.e., discrete cases) under suitable assumptions and usually formulate the structure learning problem as a discrete optimization program, while our work focuses on continuous identifiable ANMs of which the structure is fully identifiable and includes recent structure learning approaches based on continuous optimization. More discussions can be found in Appendix \ref{app:related_work}.

\section{Conclusion and Future Work}
In this paper, we propose a new approach named MissDAG to learn the underlying causal relations from incomplete data. MissDAG, leveraging the EM-based paradigm, iteratively maximizes the likelihood of the observational part of data with the inductive bias of DAG structure. Existing score-based causal discovery methods can be directly integrated into our framework for graph and model parameter learning. Moreover, MCEM is introduced to address the challenge that the closed-form posterior of missing entries is unavailable. The experiments show that our method works well across various of settings. However, our method inherits the time inefficiency issue of EM algorithm. Future works include (1) improving the sampling efficiency with more efficient sampling or variational inference techniques to approximate the posterior to scale up to larger problems, (2) incorporating other advanced causal discovery methods into the MissDAG framework, and (3) allowing unobserved confounders and cycles, e.g., using the methods by \citet{bhattacharya2021differentiable,ghassami2020characterizing}.
\section*{Acknowledgements}

LS is supported by the Major Science and Technology Innovation 2030 “Brain Science and Brain-like Research” key project (No. 2021ZD0201405). IN and KZ were partially supported by the National Institutes of Health (NIH) under Contract R01HL159805, by the NSF-Convergence Accelerator Track-D award \#2134901, by a grant from Apple Inc., and by a grant from KDDI Research Inc.. EG is supported by an Australian Government Research Training Program (RTP) Scholarship. This research was undertaken using the LIEF HPC-GPGPU Facility hosted at the University of Melbourne. This Facility was established with the assistance of LIEF Grant LE170100200. MG was supported by ARC DE210101624. TL was partially supported by Australian Research Council Projects DP180103424, DE-190101473, IC-190100031, DP-220102121, and FT-220100318. HB was supported by ARC FT190100374.

\bibliography{ref}

\begin{thebibliography}{76}
\providecommand{\natexlab}[1]{#1}
\providecommand{\url}[1]{\texttt{#1}}
\expandafter\ifx\csname urlstyle\endcsname\relax
  \providecommand{\doi}[1]{doi: #1}\else
  \providecommand{\doi}{doi: \begingroup \urlstyle{rm}\Url}\fi

\bibitem[Adel and De~Campos(2017)]{adel2017learning}
Tameem Adel and Cassio~P De~Campos.
\newblock Learning bayesian networks with incomplete data by augmentation.
\newblock In \emph{Association for the Advancement of Artificial Intelligence},
  2017.

\bibitem[Amari et~al.(1997)Amari, Chen, and Cichocki]{amari1997stability}
Shun-ichi Amari, Tian-Ping Chen, and Andrzej Cichocki.
\newblock Stability analysis of learning algorithms for blind source
  separation.
\newblock \emph{Neural Networks}, 10\penalty0 (8):\penalty0 1345--1351, 1997.

\bibitem[Andridge and Little(2010)]{andridge2010review}
Rebecca~R Andridge and Roderick~JA Little.
\newblock A review of hot deck imputation for survey non-response.
\newblock \emph{International Statistical Review}, 78\penalty0 (1):\penalty0
  40--64, 2010.

\bibitem[Bertsekas(1982)]{Bertsekas1982constrained}
Dimitri~P Bertsekas.
\newblock \emph{Constrained Optimization and {Lagrange} Multiplier Methods}.
\newblock Academic Press, 1982.

\bibitem[Bertsekas(1999)]{Bertsekas1999nonlinear}
Dimitri~P Bertsekas.
\newblock \emph{Nonlinear Programming}.
\newblock Athena Scientific, 2nd edition, 1999.

\bibitem[Bhattacharya et~al.(2020)Bhattacharya, Nabi, Shpitser, and
  Robins]{bhattacharya2020identification}
Rohit Bhattacharya, Razieh Nabi, Ilya Shpitser, and James~M Robins.
\newblock Identification in missing data models represented by directed acyclic
  graphs.
\newblock In \emph{Uncertainty in Artificial Intelligence}, pages 1149--1158.
  PMLR, 2020.

\bibitem[Bhattacharya et~al.(2021)Bhattacharya, Nagarajan, Malinsky, and
  Shpitser]{bhattacharya2021differentiable}
Rohit Bhattacharya, Tushar Nagarajan, Daniel Malinsky, and Ilya Shpitser.
\newblock Differentiable causal discovery under unmeasured confounding.
\newblock In \emph{International Conference on Artificial Intelligence and
  Statistics}, 2021.

\bibitem[Bishop(2006)]{bishop2006PRML}
Christopher~M. Bishop.
\newblock \emph{Pattern Recognition and Machine Learning}.
\newblock Springer, 2006.

\bibitem[Brouillard et~al.(2020)Brouillard, Lachapelle, Lacoste,
  Lacoste-Julien, and Drouin]{brouillard2020differentiable}
Philippe Brouillard, S{\'e}bastien Lachapelle, Alexandre Lacoste, Simon
  Lacoste-Julien, and Alexandre Drouin.
\newblock Differentiable causal discovery from interventional data.
\newblock In \emph{Advances in Neural Information Processing Systems}, 2020.

\bibitem[Cai et~al.(2011)Cai, Liu, and Luo]{Cai2011constrained}
Tony Cai, Weidong Liu, and Xi~Luo.
\newblock A constrained l1 minimization approach to sparse precision matrix
  estimation.
\newblock \emph{Journal of the American Statistical Association}, 106\penalty0
  (494):\penalty0 594--607, 2011.

\bibitem[Chen et~al.(2014)Chen, Prentice, and Wang]{chen2014penalized}
Lin~S Chen, Ross~L Prentice, and Pei Wang.
\newblock A penalized em algorithm incorporating missing data mechanism for
  gaussian parameter estimation.
\newblock \emph{Biometrics}, 70\penalty0 (2):\penalty0 312--322, 2014.

\bibitem[Chickering(2002)]{chickering2002optimal}
David~Maxwell Chickering.
\newblock Optimal structure identification with greedy search.
\newblock \emph{Journal of Machine Learning Research}, 3\penalty0
  (Nov):\penalty0 507--554, 2002.

\bibitem[Colombo et~al.(2011)Colombo, Maathuis, Kalisch, and
  Richardson]{colombo2011learning}
Diego Colombo, Marloes Maathuis, Markus Kalisch, and Thomas Richardson.
\newblock Learning high-dimensional directed acyclic graphs with latent and
  selection variables.
\newblock \emph{The Annals of Statistics}, 40:\penalty0 294--321, 2011.

\bibitem[Friedman et~al.(2008)Friedman, Hastie, and
  Tibshirani]{Friedman2008sparse}
Jerome Friedman, Trevor Hastie, and Robert Tibshirani.
\newblock Sparse inverse covariance estimation with the graphical {Lasso}.
\newblock \emph{Biostatistics}, 9:\penalty0 432--41, 2008.

\bibitem[Friedman(1997)]{friedman1997learning}
Nir Friedman.
\newblock Learning belief networks in the presence of missing values and hidden
  variables.
\newblock In \emph{International Conference on Machine Learning}, 1997.

\bibitem[Friedman(1998)]{friedman1998bayesian}
Nir Friedman.
\newblock The bayesian structural em algorithm.
\newblock In \emph{Conference on Uncertainty in Artificial Intelligence}, pages
  129--138, 1998.

\bibitem[Gain and Shpitser(2018)]{gain2018structure}
Alexander Gain and Ilya Shpitser.
\newblock Structure learning under missing data.
\newblock In \emph{International Conference on Probabilistic Graphical Models},
  pages 121--132. PMLR, 2018.

\bibitem[Geffner et~al.(2022)Geffner, Antoran, Foster, Gong, Ma, Kiciman,
  Sharma, Lamb, Kukla, Pawlowski, et~al.]{geffner2022deep}
Tomas Geffner, Javier Antoran, Adam Foster, Wenbo Gong, Chao Ma, Emre Kiciman,
  Amit Sharma, Angus Lamb, Martin Kukla, Nick Pawlowski, et~al.
\newblock Deep end-to-end causal inference.
\newblock \emph{arXiv preprint arXiv:2202.02195}, 2022.

\bibitem[Ghassami et~al.(2020)Ghassami, Yang, Kiyavash, and
  Zhang]{ghassami2020characterizing}
AmirEmad Ghassami, Alan Yang, Negar Kiyavash, and Kun Zhang.
\newblock Characterizing distribution equivalence and structure learning for
  cyclic and acyclic directed graphs.
\newblock In \emph{International Conference on Machine Learning}, 2020.

\bibitem[Ghoshal and Honorio(2018)]{ghoshal2018learning}
Asish Ghoshal and Jean Honorio.
\newblock Learning linear structural equation models in polynomial time and
  sample complexity.
\newblock In \emph{International Conference on Artificial Intelligence and
  Statistics}, pages 1466--1475. PMLR, 2018.

\bibitem[Glymour et~al.(2019)Glymour, Zhang, and Spirtes]{glymour2019review}
Clark Glymour, Kun Zhang, and Peter Spirtes.
\newblock Review of causal discovery methods based on graphical models.
\newblock \emph{Frontiers in Genetics}, 10:\penalty0 524, 2019.

\bibitem[Greenfield et~al.(2010)Greenfield, Madar, Ostrer, and
  Bonneau]{greenfield2010dream4}
Alex Greenfield, Aviv Madar, Harry Ostrer, and Richard Bonneau.
\newblock Dream4: Combining genetic and dynamic information to identify
  biological networks and dynamical models.
\newblock \emph{PloS one}, 5\penalty0 (10):\penalty0 e13397, 2010.

\bibitem[Heckerman(2008)]{heckerman2008tutorial}
David Heckerman.
\newblock A tutorial on learning with bayesian networks.
\newblock \emph{Innovations in Bayesian networks}, pages 33--82, 2008.

\bibitem[Heckerman et~al.(2006)Heckerman, Meek, and
  Cooper]{heckerman2006bayesian}
David Heckerman, Christopher Meek, and Gregory Cooper.
\newblock A bayesian approach to causal discovery.
\newblock In \emph{Innovations in Machine Learning}, pages 1--28. Springer,
  2006.

\bibitem[Hoyer et~al.(2008)Hoyer, Janzing, Mooij, Peters, and
  Sch{\"o}lkopf]{hoyer2008nonlinear}
Patrik Hoyer, Dominik Janzing, Joris~M Mooij, Jonas Peters, and Bernhard
  Sch{\"o}lkopf.
\newblock Nonlinear causal discovery with additive noise models.
\newblock In \emph{Advances in Neural Information Processing Systems}, 2008.

\bibitem[Hyv{\"a}rinen et~al.(2010)Hyv{\"a}rinen, Zhang, Shimizu, and
  Hoyer]{hyvarinen2010estimation}
Aapo Hyv{\"a}rinen, Kun Zhang, Shohei Shimizu, and Patrik~O Hoyer.
\newblock Estimation of a structural vector autoregression model using
  non-gaussianity.
\newblock \emph{Journal of Machine Learning Research}, 11\penalty0 (5), 2010.

\bibitem[Imbens and Rubin(2015)]{imbens2015causal}
Guido~W Imbens and Donald~B Rubin.
\newblock \emph{Causal inference in statistics, social, and biomedical
  sciences}.
\newblock Cambridge University Press, 2015.

\bibitem[Koller and Friedman(2009)]{koller2009probabilistic}
Daphne Koller and Nir Friedman.
\newblock \emph{Probabilistic graphical models: principles and techniques}.
\newblock MIT press, 2009.

\bibitem[Kyono et~al.(2021)Kyono, Zhang, Bellot, and van~der
  Schaar]{kyono2021miracle}
Trent Kyono, Yao Zhang, Alexis Bellot, and Mihaela van~der Schaar.
\newblock Miracle: Causally-aware imputation via learning missing data
  mechanisms.
\newblock In \emph{Advances in Neural Information Processing Systems},
  volume~34, 2021.

\bibitem[Lachapelle et~al.(2020)Lachapelle, Brouillard, Deleu, and
  Lacoste-Julien]{lachapelle2020gradient}
S{\'e}bastien Lachapelle, Philippe Brouillard, Tristan Deleu, and Simon
  Lacoste-Julien.
\newblock Gradient-based neural dag learning.
\newblock In \emph{International Conference on Learning Representations}, 2020.

\bibitem[Little and Rubin(2019)]{little2019statistical}
Roderick~JA Little and Donald~B Rubin.
\newblock \emph{Statistical analysis with missing data}, volume 793.
\newblock John Wiley \& Sons, 2019.

\bibitem[Ma and Zhang(2021)]{ma2021identifiable}
Chao Ma and Cheng Zhang.
\newblock Identifiable generative models for missing not at random data
  imputation.
\newblock \emph{Advances in Neural Information Processing Systems},
  34:\penalty0 27645--27658, 2021.

\bibitem[Mayer et~al.(2020)Mayer, Josse, Raimundo, and
  Vert]{mayer2020missdeepcausal}
Imke Mayer, Julie Josse, F{\'e}lix Raimundo, and Jean-Philippe Vert.
\newblock Missdeepcausal: Causal inference from incomplete data using deep
  latent variable models.
\newblock \emph{arXiv preprint arXiv:2002.10837}, 2020.

\bibitem[Mohan and Pearl(2021)]{mohan2021graphical}
Karthika Mohan and Judea Pearl.
\newblock Graphical models for processing missing data.
\newblock \emph{Journal of the American Statistical Association}, pages 1--16,
  2021.

\bibitem[Mooij et~al.(2011)Mooij, Janzing, Heskes, and
  Sch{\"o}lkopf]{mooij2011causal}
Joris~M Mooij, Dominik Janzing, Tom Heskes, and Bernhard Sch{\"o}lkopf.
\newblock On causal discovery with cyclic additive noise model.
\newblock In \emph{Advances in Neural Information Processing Systems}, 2011.

\bibitem[Muzellec et~al.(2020)Muzellec, Josse, Boyer, and
  Cuturi]{muzellec2020missing}
Boris Muzellec, Julie Josse, Claire Boyer, and Marco Cuturi.
\newblock Missing data imputation using optimal transport.
\newblock In \emph{International Conference on Machine Learning}, pages
  7130--7140. PMLR, 2020.

\bibitem[Nabi et~al.(2020)Nabi, Bhattacharya, and Shpitser]{nabi2020full}
Razieh Nabi, Rohit Bhattacharya, and Ilya Shpitser.
\newblock Full law identification in graphical models of missing data:
  Completeness results.
\newblock In \emph{International Conference on Machine Learning}, pages
  7153--7163. PMLR, 2020.

\bibitem[Neath(2013)]{neath2013convergence}
Ronald~C Neath.
\newblock On convergence properties of the monte carlo em algorithm.
\newblock \emph{Advances in Modern Statistical Theory and Applications: a
  Festschrift in Honor of Morris L. Eaton}, pages 43--62, 2013.

\bibitem[Ng et~al.(2020)Ng, Ghassami, and Zhang]{ng2020role}
Ignavier Ng, AmirEmad Ghassami, and Kun Zhang.
\newblock On the role of sparsity and dag constraints for learning linear dags.
\newblock In \emph{Advances in Neural Information Processing Systems},
  volume~33, 2020.

\bibitem[Ng et~al.(2022)Ng, Zhu, Fang, Li, Chen, and Wang]{ng2019masked}
Ignavier Ng, Shengyu Zhu, Zhuangyan Fang, Haoyang Li, Zhitang Chen, and Jun
  Wang.
\newblock Masked gradient-based causal structure learning.
\newblock In \emph{SIAM International Conference on Data Mining}, pages
  424--432, 2022.

\bibitem[Pamfil et~al.(2020)Pamfil, Sriwattanaworachai, Desai, Pilgerstorfer,
  Georgatzis, Beaumont, and Aragam]{pamfil2020dynotears}
Roxana Pamfil, Nisara Sriwattanaworachai, Shaan Desai, Philip Pilgerstorfer,
  Konstantinos Georgatzis, Paul Beaumont, and Bryon Aragam.
\newblock {Dynotears: Structure learning from time-series data}.
\newblock In \emph{International Conference on Artificial Intelligence and
  Statistics}, 2020.

\bibitem[Pearl(2009)]{pearl2009causality}
Judea Pearl.
\newblock \emph{Causality: Models, reasoning, and inference}.
\newblock Cambridge University Press, 2009.

\bibitem[Pearl et~al.(2016)Pearl, Glymour, and Jewell]{pearl2016causal}
Judea Pearl, Madelyn Glymour, and Nicholas~P Jewell.
\newblock \emph{Causal inference in statistics: A primer}.
\newblock John Wiley \& Sons, 2016.

\bibitem[Peters et~al.(2014)Peters, Mooij, Janzing, and
  Sch{\"o}lkopf]{peters2014causal}
J~Peters, JM~Mooij, D~Janzing, and B~Sch{\"o}lkopf.
\newblock Causal discovery with continuous additive noise models.
\newblock \emph{Journal of Machine Learning Research}, 15\penalty0
  (1):\penalty0 2009--2053, 2014.

\bibitem[Peters and B{\"u}hlmann(2014)]{peters2014identifiability}
Jonas Peters and Peter B{\"u}hlmann.
\newblock Identifiability of gaussian structural equation models with equal
  error variances.
\newblock \emph{Biometrika}, 101\penalty0 (1):\penalty0 219--228, 2014.

\bibitem[Raskutti and Uhler(2018)]{raskutti2018learning}
Garvesh Raskutti and Caroline Uhler.
\newblock Learning directed acyclic graph models based on sparsest
  permutations.
\newblock \emph{Stat}, 7\penalty0 (1):\penalty0 e183, 2018.

\bibitem[Resnik(2008)]{resnik2008randomized}
David~B Resnik.
\newblock Randomized controlled trials in environmental health research:
  ethical issues.
\newblock \emph{Journal of Environmental Health}, 70\penalty0 (6):\penalty0 28,
  2008.

\bibitem[Richens et~al.(2020)Richens, Lee, and Johri]{richens2020improving}
Jonathan~G Richens, Ciar{\'a}n~M Lee, and Saurabh Johri.
\newblock Improving the accuracy of medical diagnosis with causal machine
  learning.
\newblock \emph{Nature communications}, 11\penalty0 (1):\penalty0 1--9, 2020.

\bibitem[Riggelsen(2006)]{riggelsen2006learning}
Carsten Riggelsen.
\newblock Learning bayesian networks from incomplete data: An efficient method
  for generating approximate predictive distributions.
\newblock In \emph{SIAM International Conference on Data Mining}, pages
  130--140. SIAM, 2006.

\bibitem[Robert et~al.(2004)Robert, Casella, and Casella]{robert2004monte}
Christian~P Robert, George Casella, and George Casella.
\newblock \emph{Monte Carlo statistical methods}, volume~2.
\newblock Springer, 2004.

\bibitem[Rubin(1976)]{rubin1976inference}
Donald~B Rubin.
\newblock Inference and missing data.
\newblock \emph{Biometrika}, 63\penalty0 (3):\penalty0 581--592, 1976.

\bibitem[Schwarz(1978)]{Schwarz1978estimating}
Gideon Schwarz.
\newblock Estimating the dimension of a model.
\newblock \emph{The Annals of Statistics}, pages 461--464, 1978.

\bibitem[Shimizu et~al.(2006)Shimizu, Hoyer, Hyv{\"a}rinen, Kerminen, and
  Jordan]{shimizu2006linear}
Shohei Shimizu, Patrik~O Hoyer, Aapo Hyv{\"a}rinen, Antti Kerminen, and Michael
  Jordan.
\newblock A linear non-gaussian acyclic model for causal discovery.
\newblock \emph{Journal of Machine Learning Research}, 7\penalty0 (10), 2006.

\bibitem[Shimizu et~al.(2011)Shimizu, Inazumi, Sogawa, Hyv{\"a}rinen, Kawahara,
  Washio, Hoyer, and Bollen]{Shimizu2011directlingam}
Shohei Shimizu, Takanori Inazumi, Yasuhiro Sogawa, Aapo Hyv{\"a}rinen,
  Yoshinobu Kawahara, Takashi Washio, Patrik~O Hoyer, and Kenneth Bollen.
\newblock {DirectLiNGAM}: A direct method for learning a linear {non-Gaussian}
  structural equation model.
\newblock \emph{Journal of Machine Learning Research}, 12\penalty0
  (Apr):\penalty0 1225--1248, 2011.

\bibitem[Shpitser et~al.(2014)Shpitser, Evans, Richardson, and
  Robins]{shpitser2014introduction}
Ilya Shpitser, Robin~J Evans, Thomas~S Richardson, and James~M Robins.
\newblock Introduction to nested markov models.
\newblock \emph{Behaviormetrika}, 41\penalty0 (1):\penalty0 3--39, 2014.

\bibitem[Singh(1997)]{singh1997learning}
Moninder Singh.
\newblock Learning bayesian networks from incomplete data.
\newblock In \emph{Association for the Advancement of Artificial Intelligence},
  pages 534--539, 1997.

\bibitem[Spirtes and Glymour(1991)]{spirtes1991pc}
Peter Spirtes and Clark Glymour.
\newblock An algorithm for fast recovery of sparse causal graphs.
\newblock \emph{Social Science Computer Review}, 9:\penalty0 62--72, 1991.

\bibitem[Spirtes et~al.(2001)Spirtes, Glymour, Scheines,
  et~al.]{spirtes2001causation}
Peter Spirtes, Clark Glymour, Richard Scheines, et~al.
\newblock \emph{Causation, Prediction, and Search}, volume~1.
\newblock The MIT Press, 2001.

\bibitem[St{\"a}dler and B{\"u}hlmann(2012)]{stadler2012missing}
Nicolas St{\"a}dler and Peter B{\"u}hlmann.
\newblock Missing values: sparse inverse covariance estimation and an extension
  to sparse regression.
\newblock \emph{Statistics and Computing}, 22\penalty0 (1):\penalty0 219--235,
  2012.

\bibitem[Stekhoven and B{\"u}hlmann(2012)]{stekhoven2012missforest}
Daniel~J Stekhoven and Peter B{\"u}hlmann.
\newblock Missforest—non-parametric missing value imputation for mixed-type
  data.
\newblock \emph{Bioinformatics}, 28\penalty0 (1):\penalty0 112--118, 2012.

\bibitem[Tanner and Wong(1987)]{tanner1987calculation}
Martin~A Tanner and Wing~Hung Wong.
\newblock The calculation of posterior distributions by data augmentation.
\newblock \emph{Journal of the American statistical Association}, 82\penalty0
  (398):\penalty0 528--540, 1987.

\bibitem[Tu et~al.(2019)Tu, Zhang, Ackermann, Mohan, Kjellstr{\"o}m, and
  Zhang]{tu2019causal}
Ruibo Tu, Cheng Zhang, Paul Ackermann, Karthika Mohan, Hedvig Kjellstr{\"o}m,
  and Kun Zhang.
\newblock Causal discovery in the presence of missing data.
\newblock In \emph{International Conference on Artificial Intelligence and
  Statistics}, pages 1762--1770. PMLR, 2019.

\bibitem[Wang et~al.(2021)Wang, Du, Zhu, Ke, Chen, Hao, and
  Wang]{wang2021ordering}
Xiaoqiang Wang, Yali Du, Shengyu Zhu, Liangjun Ke, Zhitang Chen, Jianye Hao,
  and Jun Wang.
\newblock Ordering-based causal discovery with reinforcement learning.
\newblock In \emph{International Joint Conference on Artificial Intelligence},
  2021.

\bibitem[Wang et~al.(2020)Wang, Liang, Charlin, and Blei]{wang2020causal}
Yixin Wang, Dawen Liang, Laurent Charlin, and David~M Blei.
\newblock Causal inference for recommender systems.
\newblock In \emph{ACM Conference on Recommender Systems}, pages 426--431,
  2020.

\bibitem[Wei and Tanner(1990)]{wei1990monte}
Greg~CG Wei and Martin~A Tanner.
\newblock A monte carlo implementation of the em algorithm and the poor man's
  data augmentation algorithms.
\newblock \emph{Journal of the American statistical Association}, 85\penalty0
  (411):\penalty0 699--704, 1990.

\bibitem[White et~al.(2011)White, Royston, and Wood]{white2011multiple}
Ian~R White, Patrick Royston, and Angela~M Wood.
\newblock Multiple imputation using chained equations: issues and guidance for
  practice.
\newblock \emph{Statistics in medicine}, 30\penalty0 (4):\penalty0 377--399,
  2011.

\bibitem[Wu(1983)]{wu1983convergence}
CF~Jeff Wu.
\newblock On the convergence properties of the em algorithm.
\newblock \emph{The Annals of Statistics}, pages 95--103, 1983.

\bibitem[Yoon et~al.(2018)Yoon, Jordon, and Schaar]{yoon2018gain}
Jinsung Yoon, James Jordon, and Mihaela Schaar.
\newblock Gain: Missing data imputation using generative adversarial nets.
\newblock In \emph{International Conference on Machine Learning}, pages
  5689--5698. PMLR, 2018.

\bibitem[Yu et~al.(2019)Yu, Chen, Gao, and Yu]{yu2019dag}
Yue Yu, Jie Chen, Tian Gao, and Mo~Yu.
\newblock {DAG-GNN: DAG structure learning with graph neural networks}.
\newblock In \emph{International Conference on Machine Learning}, 2019.

\bibitem[Yuan and Malone(2013)]{Yuan2013learning}
Changhe Yuan and Brandon Malone.
\newblock Learning optimal {Bayesian} networks: A shortest path perspective.
\newblock \emph{Journal of Artificial Intelligence Research}, 48\penalty0
  (1):\penalty0 23--65, 2013.

\bibitem[Yuan et~al.(2011)Yuan, Malone, and Wu]{yuan2011learning}
Changhe Yuan, Brandon Malone, and Xiaojian Wu.
\newblock Learning optimal bayesian networks using a* search.
\newblock In \emph{International Joint Conference on Artificial Intelligence},
  2011.

\bibitem[Zhang and Hyvarinen(2009)]{zhang2009identifiability}
Kun Zhang and Aapo Hyvarinen.
\newblock On the identifiability of the post-nonlinear causal model.
\newblock In \emph{Conference on Uncertainty in Artificial Intelligence}, 2009.

\bibitem[Zheng(2020)]{zheng2020thsis}
Xun Zheng.
\newblock \emph{Learning DAGs with Continuous Optimization}.
\newblock PhD thesis, Carnegie Mellon University, 2020.

\bibitem[Zheng et~al.(2018)Zheng, Aragam, Ravikumar, and Xing]{zheng2018dags}
Xun Zheng, Bryon Aragam, Pradeep~K Ravikumar, and Eric~P Xing.
\newblock {DAGs with NO TEARS: Continuous Optimization for Structure Learning}.
\newblock In \emph{Advances in Neural Information Processing Systems},
  volume~31, 2018.

\bibitem[Zheng et~al.(2020)Zheng, Dan, Aragam, Ravikumar, and
  Xing]{zheng2020learning}
Xun Zheng, Chen Dan, Bryon Aragam, Pradeep Ravikumar, and Eric~P. Xing.
\newblock Learning sparse nonparametric {DAGs}.
\newblock In \emph{International Conference on Artificial Intelligence and
  Statistics}, 2020.

\bibitem[Zhu et~al.(2020)Zhu, Ng, and Chen]{zhu2020causal}
Shengyu Zhu, Ignavier Ng, and Zhitang Chen.
\newblock Causal discovery with reinforcement learning.
\newblock In \emph{International Conference on Learning Representations}, 2020.

\end{thebibliography}
\bibliographystyle{plainnat}

\section*{Checklist}

\begin{enumerate}

\item For all authors...
\begin{enumerate}
  \item Do the main claims made in the abstract and introduction accurately reflect the paper's contributions and scope?
    \answerYes{}
  \item Did you describe the limitations of your work?
    \answerYes{}
  \item Did you discuss any potential negative societal impacts of your work?
    \answerNA{}
  \item Have you read the ethics review guidelines and ensured that your paper conforms to them?
    \answerYes{}
\end{enumerate}

\item If you are including theoretical results...
\begin{enumerate}
  \item Did you state the full set of assumptions of all theoretical results?
    \answerYes{}
        \item Did you include complete proofs of all theoretical results?
    \answerYes{}
\end{enumerate}

\item If you ran experiments...
\begin{enumerate}
  \item Did you include the code, data, and instructions needed to reproduce the main experimental results (either in the supplemental material or as a URL)?
    \answerYes{}
  \item Did you specify all the training details (e.g., data splits, hyperparameters, how they were chosen)?
    \answerYes{}
        \item Did you report error bars (e.g., with respect to the random seed after running experiments multiple times)?
    \answerYes{}
        \item Did you include the total amount of compute and the type of resources used (e.g., type of GPUs, internal cluster, or cloud provider)?
    \answerYes{}
\end{enumerate}

\item If you are using existing assets (e.g., code, data, models) or curating/releasing new assets...
\begin{enumerate}
  \item If your work uses existing assets, did you cite the creators?
    \answerYes{}
  \item Did you mention the license of the assets?
    \answerYes{}
  \item Did you include any new assets either in the supplemental material or as a URL?
    \answerNo{}
  \item Did you discuss whether and how consent was obtained from people whose data you're using/curating?
    \answerNo{}
  \item Did you discuss whether the data you are using/curating contains personally identifiable information or offensive content?
    \answerNA{}
\end{enumerate}

\item If you used crowdsourcing or conducted research with human subjects...
\begin{enumerate}
  \item Did you include the full text of instructions given to participants and screenshots, if applicable?
    \answerNA{}
  \item Did you describe any potential participant risks, with links to Institutional Review Board (IRB) approvals, if applicable?
    \answerNA{}
  \item Did you include the estimated hourly wage paid to participants and the total amount spent on participant compensation?
    \answerNA{}
\end{enumerate}

\end{enumerate}

\clearpage
\appendix
\onecolumn
{\LARGE \centerline{\textbf{Supplementary materials}}}
\section{Additional related works}
\label{app:related_work}

\textbf{Comparison with \citet{friedman1997learning,singh1997learning}.} \ 
Our work shares similarities with [15, 55], since both rely on the EM algorithm. However, our work focuses on continuous identifiable ANMs that have recently received considerable attention \citep{peters2014causal}, while \citet{friedman1997learning,singh1997learning} focus on discrete cases in which one is only able to identify the Markov equivalence class; therefore, the key technical development is different. (1) For the linear Gaussian case, we derive the closed-form solution of exact posterior that is different from the discrete case considered by \citet{friedman1997learning,singh1997learning}. (2) For the linear non-Gaussian and nonlinear cases, since the exact posterior is not available in closed form, we develop a method based on approximate posterior using Monte Carlo and rejection sampling; such a setup that involves approximate posterior may be more challenging and has not been considered by \citet{friedman1997learning,singh1997learning}, since the exact posterior of discrete case considered by \citet{friedman1997learning,singh1997learning} is available in closed form. (3) Our formulation includes modern structure learning approaches based on continuous optimization (in addition to classical methods based on discrete optimization considered by \citet{friedman1997learning,singh1997learning}).

\section{Proofs}
\label{app:proofs}

\subsection{Proof of Proposition~\ref{lemma:det_equals_1}}\label{app:proofs_det_equals_1}
We take $B_{\theta_{\mathcal{G}}}$ as the adjacency matrix of $\theta_{\mathcal{G}}$. If $({B_{\theta_{\mathcal{G}}}})_{ij} = 0$, then, $({\mathbf{J}_{\theta^t_f}})_{ij}=0$, i.e., $\mathbf{J}_{\theta^t_f}$ implicitly encodes a DAG structure. Therefore, there exists a permutation matrix $U$ such that $U\mathbf{J}_{\theta^t_f}U^T$ is strictly upper triangular. Then, $|\det (\mathbf{I} - \mathbf{J}_{\theta^t_f})|=|\det (\mathbf{I} - U\mathbf{J}_{\theta^t_f}U^T)|=1$.
\section{Implementation details}
\label{app:implementation_details}
We provide the implementation details for the structure learning and imputation methods, and for the procedure used to generate the missing data. We also describe the hyperparameters used for the proposed MissDAG framework.

\subsection{Structure learning methods}\label{app:structure_learning_methods}
We use existing implementations for most structure learning methods:
\begin{itemize}[leftmargin=*]
\item \textbf{A* and GES}\footnote{\label{note:causal_learn}\url{https://github.com/cmu-phil/causal-learn}}: A* \citep{yuan2011learning,Yuan2013learning} formulates the score-based structure learning problem as a shortest path problem and uses the A* search procedure with a consistent heuristic function to guide the search in the search space of DAGs, and is guaranteed to return the optimal DAG. On the other hand, GES \citep{chickering2002optimal} adopts a greedy search procedure in the search space of equivalence classes. Therefore, in the M-step, one has to convert the estimated equivalence class by GES into a consistent DAG. For both methods, we adopt the BIC score \citep{Schwarz1978estimating}. 
\item \textbf{Testwise Deletion PC (TD-PC)}: TD-PC \citep{tu2019causal} is an extension of PC that makes use of all instances without any missing value for the variables involved in the conditional independence test.
It provides asymptotically correct results for the MCAR case while may not give correct for the MAR case, since the condition $p(R|X_m, X_o)=p(R)$ does not hold. Here we use the Fisher-z test and set the $p$-value to $0.05$.
\item \textbf{NOTEARS, NOTEARS-ICA, and NOTEARS-MLP}\footnote{\url{https://github.com/xunzheng/notears}}: NOTEARS-based methods are widely used in our paper, including NOTEARS \citep{zheng2018dags} for the LGM, NOTEAES-ICA (NOTEARS-ICA-MCEM)  \citep{zheng2020thsis} for the LiNGAM, and NOTEARS-MLP (NOTEARS-MLP-MCEM) \citep{zheng2020learning} for the NL-ANMs. For all NOTEARS-based models, we follow the original papers and use the augmented Lagrangian method to solve the constrained optimization problem; see Appendix \ref{app:alm} for further details. We adopt the same hyperparameters suggested in the original papers, since we do not aim to report the best performance for all settings by carefully tuning these parameters. In particular, the initial $\alpha$ and $\rho$ are both set to $0$. The other hyperparameters $\gamma$, $h_{min}$ and $\rho_{max}$ are set to $0.25$, $1\times10^{-8}$ and $1\times 10^{16}$, respectively.
The sparsity parameter $\lambda_1$ for NOTEARS, NOTEARS-ICA (NOTEARS-ICA-MCEM), NOTEARS-MLP (NOTEARS-MLP-MCEM) is set to $0.1$, $0.1$, and $0.03$, respectively. Moreover, NOTEARS-MLP (NOTEARS-MLP-MCEM) also applies a $\ell_2$ penalty to all the weights of the multilayer perceptrons, whose coefficient $\lambda_2$ is set to $0.01$.
\item \textbf{GOLEM}\footnote{\url{https://github.com/ignavierng/golem}}: \citet{ng2020role} show that, when likelihood-based objective is used together with the soft sparsity and DAG constraints, it is able to to recover the true structure under certain conditions. They further proposed an algorithm, called GOLEM, to do so, which involves solving an unconstrained optimization problem. The hyperparameters need for GOLEM is just (1) $\lambda_1$ for the $\ell_1$ sparsity and (2) $\lambda_2$ for the DAG constraint penalty. Throughout all experiments, we set them to $\lambda_1=5\times 10^{-2}$ and $\lambda_2=5\times 10^{-3}$, which are slightly larger than the original ones used by \citet{ng2020role} and found to be more effective when the sample size is small. In the NOTEARS-ICA-MCEM and MOTEARS-MLP-MCEM, to construct $q(X_{\mathbf{m}})$, we first use zero imputation to impute $\mathbf{X}_{\mathbf{O}}$ and get $\hat{\mathbf{X}}$. For each observation $\mathbf{X}_i$ and $j\in \mathbf{m}$, we calculate $\sigma_j^2$ from $\hat{\mathbf{X}}$. Then, a diag matrix is constructed as $\Sigma = \text{diag}(\sigma_j^2:j\in \mathbf{m})$ to be a covariance matrix. Finally, $q(X_{\mathbf{m}})$ is defined as a multivariate Gaussian distribution with zero-mean and $\Sigma$ as the covariance matrix.
\item \textbf{ICA-LiNGAM and Direct-LiNGAM}\footnote{\url{https://github.com/cdt15/lingam}}: ICA-LiNGAM \citep{shimizu2006linear} utilizes independent component analysis to estimate the LiNGAM, while Direct-LiNGAM \citep{Shimizu2011directlingam} recovers the causal order of the variables by iteratively removing the effect of each variable from the data.
\item \textbf{Ghoshal}: The algorithm described by \citet{ghoshal2018learning} first estimates the inverse covariance matrix, and then iteratively identifies and removes a terminal node. The parent set and edge weights are also estimated during the iterative procedure. We adopt our own implementation of the algorithm  because we did not manage to find a publicly available implementation. The original algorithm employs the CLIME method \citep{Cai2011constrained} to estimate the inverse covariance matrix, while we use the graphical Lasso method \citep{Friedman2008sparse}.
\end{itemize}
Note that the experiments for GOLEM are conducted on NVIDIA V100 GPU, while those for the other methods are conducted on CPU instances.


\subsection{Imputation methods}
We use existing implementations for most imputation methods:
\begin{itemize}[leftmargin=*]
\item \textbf{MissForest imputation and KNNImputer}\footnote{\url{https://github.com/epsilon-machine/missingpy}}: The default hyperparameters are used.
\item \textbf{MICE imputation}\footnote{\url{https://github.com/scikit-learn/scikit-learn}}: 
We set the hyperparameter \textit{n-imputations} as $1$.
\item \textbf{GAIN and optimal transport (OT) imputation}\footnote{\url{https://github.com/trentkyono/MIRACLE}}: GAIN is an adversarial learning framework that consists of two generators, which are used to impute the missing entries and generate the hint matrix, respectively, and of a discriminator that is used to distinguish between observed and imputed entries. The hyperparameter $\alpha$ is set to $10$ and the hint rate is set to $0.9$. The learning rate is taken as $1\times 10^{-3}$ and there are a total of $10000$ iterations for the adversarial learning procedure. OT imputation leverages the optimal transport distances and integrate it into the loss functions to achieve the imputation. In OT imputation, we set the learning rate and $\epsilon$ to $0.01$, the number of iterations to $500$, and the scaling parameter in Sinkhorn iterations to $0.9$.
\item \textbf{Mean and Gaussian-EM imputation}: We adopt our own implementation of these two imputation methods. Mean imputation fills the missing entries using the average of the observed values of the corresponding variable. For Gaussian-EM imputation, the E-step is the same as our method, while in the M-step, the estimated statistic $\bf T$ is directly return to the E-step without any further operation.
There is no extra hyperparameter used for these two method.
\end{itemize}


\subsection{Missingness}
According to the underlying reasons why the data are missing, the missing mechanisms are typically classified into three categories. (1) MCAR. The missing mechanism is independent from all variables. (2) MAR. The missing mechanism is systematically related to the observational variables but independent from the missing variables. (3) MNAR. The missing mechanism is related to the missing variables.
We describe the procedure to generate the missing matrix $\bf Y$, which is used to mask the synthetic data $\bf X$ to simulate different types of missing data, i.e., MCAR, MAR, and MNAR with missing rate $r_m$.
\begin{itemize}[leftmargin=*]
\item \textbf{MCAR}: Firstly, sampling a matrix $\bf Y'$ from a $\operatorname{Uniform} ([0,1])$, and set $\mathbf{Y}_{ij} = 0$ if $\mathbf{Y'}_{ij} \leq r_m$ and $\mathbf{Y}_{ij} = 1$ otherwise. For the experiment in Appendix \ref{app:different_missing_types} that compare against different missing types, we specifically set $30\%$ of the variables to be full-observational, i.e., without any missing value, in order to ensure a relatively fair comparison with the MAR case.
\item \textbf{MAR}: We set $30\%$ of the variables to be fully-observed. Then, the missingness of the remaining variables are generated according to a logistic model with random weights that are related to the fully observed variables.
\item  \textbf{MNAR}: The self-masked missingness is taken as the MNAR mechanism. To ensure a relatively fair comparison with the MAR case, $30\%$ of the variables do not have any missing value. Then, the remaining variables are masked according to a logistic model with random weights that are related to the corresponding variables.
\end{itemize}

\subsection{Hyperparameters of MissDAG}
The proposed MissDAG framework is able to solve four types of ANMs, including LGM-EV, LGM-NV, LiNGAM, and NL-ANM. For different models, different causal discovery or structure learning methods are leveraged, each of which involves a different set of hyperparameters, described in Appendix \ref{app:structure_learning_methods}. To ensure a fair comparison, we use the same set of hyperparameters for these structure learning methods across different imputation methods, and our MissDAG framework.
Our framework involves an additional hyperparameter corresponding to the number of iterations for the EM procedure, which we set to $10$.



\section{Solving the optimization problem}
\label{app:solving_optimization}

\subsection{Augmented Lagrangian method}\label{app:alm}
Here, we rewrite the equality constrained optimization problem in the M-step (here, we equivalently minimize the negative score function.) of MissDAG as follows:
\begin{equation*}
\begin{aligned}
   \arg \min_{\theta} & \ \ -\mathcal{S}(\theta) + \lambda\ \textbf{PEN}(\theta), \\
    \text{subject to} & \ \ \mathcal{\theta}_{\mathcal{G}} \in \textbf{DAGs}, \iff h(\theta_{\mathcal{G}}) = \text{Tr}(\exp(\theta_{\mathcal{G}})) -d = 0,
\end{aligned}
\end{equation*}
where $\exp(\cdot)$ is the matrix exponential and $\text{Tr}(\cdot)$ calculates the matrix trace. In NOTEARS~\citep{zheng2018dags}, the above optimization problem is solved by leveraging the augmented Lagrangian method \citep{Bertsekas1982constrained,Bertsekas1999nonlinear} to get an approximate solution. It is an iterative-based optimization method, which transforms the optimization object into a series of unconstrained sub-problems. The $t$-th sub-problem involving the augmented Lagrangian can be formulated as 
\begin{equation*}
    \arg \min_{\theta} \ \  -\mathcal{S}(\theta) + \lambda\ \textbf{PEN}(\theta) + \alpha_t h(\theta_{\mathcal{G}}) + \frac{\rho_t}{2} |h(\theta_{\mathcal{G}})|^2,
\end{equation*}
where $\alpha_t$ and $\rho_t$ are the parameters updated by the iterative step, which represent the estimate of the Lagrange Multiplier and the penalty parameter, respectively. The values of these two parameters are gradually increased to make the final solution approximately meet the requirement of equality constraint. Specifically, the iterative step follows the following update rules:
\begin{equation*}
\begin{aligned}
    \theta^{t+1} &= \argmin_{\theta} \ \ -\mathcal{S}(\theta) + \lambda\ \textbf{PEN}(\theta) + \alpha_t h(\theta_{\mathcal{G}}) + \frac{\rho_t}{2} |h(\theta_{\mathcal{G}})|^2, \\
    \alpha^{t+1} &= \alpha^t + \rho^t  h(\theta_{\mathcal{G}}^t) \\
    \rho^{t+1} &= \begin{cases}
    \beta  \rho^t & \text{if } h(\theta_{\mathcal{G}}^{t+1})  \geq \gamma h(\theta_{\mathcal{G}}^t), \\
    \rho^t & \text{otherwise},
    \end{cases}
\end{aligned}
\end{equation*}
where $\beta$ and $\gamma$ are the hyperparameters. For NOTEARS-ICA and NOTEARS-MLP, we also use the augmented Lagrangian method to solve the problem in which the only difference is the score function.

\subsection{Soft constraints}\label{app:soft_constraints}
GOLEM~\citep{ng2020role} employs likelihood-based objective with soft sparsity and DAG constraints for structure learning. However, in our setting of missing data, NOTEARS fits better into our MissDAG framework, specifically in the M-step, as compared to GOLEM, as the former solves a constrained optimization problem and is guaranteed to return DAGs. However, this does not lead to the conclusion that the least squares loss used by NOTEARS is better than the likelihood-based objective used by GOLEM. This is because the study by \citet{ng2020role} shows that under certain conditions, likelihood-based objective with the soft constraints introduced are able to to recover the true structure. In other words, unlike NOTEARS, we do not have to enforce a hard acyclicity constraint, and the unconstrained optimization problem will return a solution close to being a DAG in practice. The optimization problem is as follows,
\begin{equation*}
    \argmin_{\theta_{\mathcal{M}}}  -\mathcal{S}_{\text{GOLEM}} + \lambda_1 \ \textbf{PEN}(\theta_{\mathcal{M}}) + \lambda_2 h(\theta_{\mathcal{G}}).
\end{equation*}
where $\lambda_1$ and $\lambda_2$ are the penalty coefficients. With the likelihood-based objective, GOLEM can also be applied to the non-identifiable LGM-NV case to identify the Markov equivalence class of the ground-truth DAG.

\subsection{Thresholding}
As suggested by \citet{zheng2020thsis,ng2020role}, the solutions produced by the methods described in Appendices \ref{app:alm} and \ref{app:soft_constraints} usually contain a number of entries with a small magnitude; therefore,
thresholding is used to alleviate this problem. For these methods, we follow the original papers and set the threshold to $0.3$ to prune the learned adjacency matrix to get the final graph. To guarantee the DAG output, an iterative deletion method is also taken, which cut off the edge with the minimum magnitude until obtaining a DAG. Note that we only apply this post-processing step after the EM procedure ends to obtain the final graph, but not during the EM procedure.

\section{Convergence analysis}
\label{app:convergence_analysis}

The convergence of MissDAG is highly relied on the convergence properties of EM~\citep{wu1983convergence} and MCEM framework~\citep{neath2013convergence}. Moreover, the convergence analysis of Bayesian network learning from incomplete data has been well provided by \citet{stadler2012missing, friedman1998bayesian}. For completeness, we provide similar conclusions in this section that meet the different cases in our framework.

To prove the convergence of MissDAG, a penalized EM-based iterative method, we can turn to prove the establishment of $p(X_o; \theta^{t+1}) \geq p(X_o; \theta^{t})$ or, equally, $\log p(X_o; \theta^{t+1}) \geq \log p(X_o; \theta^{t})$. With Bayes Rule, we have 
\begin{equation*}
    \log p(X_o; \theta) = \log p(X_o, X_m; \theta) - \log p(X_m|X_o; \theta).
\end{equation*}
And then, we get the expectation over the missing variables given the observed variables and the parameters $\theta_t$ on both sides of the equation to obtain
\begin{align*}
  \mathbb{E}_{X_m|X_o; \theta^{t}} \{ \log p(X_o; \theta) \} =& \mathbb{E}_{X_m|X_o; \theta^{t}} \{\log p(X_o, X_m; \theta)\} -\mathbb{E}_{X_m|X_o; \theta^{t}} \{\log p(X_m |X_o; \theta)\} \\
  \Rightarrow	 \log p(X_o; \theta)  =& \mathbb{E}_{X_m|X_o; \theta^{t}} \{\log p(X_o, X_m; \theta)\} - \mathbb{E}_{X_m|X_o; \theta^{t}} \{\log p(X_m |X_o; \theta)\}.
\end{align*}
The first term of the RHS with sparsity constraint corresponds to the term $\mathcal{Q}(\theta,\theta^{t})$ in our paper. Since $\theta^{t+1}=\arg \max_{\theta}\mathcal{Q}(\theta,\theta^{t})$, we have
\begin{equation}
    \mathcal{Q}(\theta^{t+1},\theta^{t}) \geq \mathcal{Q}(\theta^{t},\theta^{t}).
    \label{eq:Q-inequality}
\end{equation}
Furthermore, to prove $\log p(X_o; \theta^{t+1}) \geq \log p(X_o; \theta^{t})$, we also need 
\begin{equation*}
    \mathbb{E}_{X_m |X_o; \theta^{t}} \{\log P(X_m|X_o; \theta^{t+1})\} \leq \mathbb{E}_{X_m|X_o; \theta^{t}} \{\log P(X_{m}|X_o; \theta^{t})\},
\end{equation*}
or, equivalently, 
\begin{equation*}
    \mathbb{E}_{X_m|X_o; \theta^{t}} \left\{\log \frac{p(X_m |X_o; \theta^{t+1})}{ p(X_m|X_o; \theta^{t})} \right\} \leq 0.
\end{equation*}
This is straightforward to be proved since
\begin{equation*}
\begin{aligned}
     \mathbb{E}_{X_m|X_o; \theta^{t}} \left\{\log \frac{p(X_m |X_o; \theta^{t+1})}{ p(X_m|X_o; \theta^{t})} \right\} &= - D_{KL}\left(p(X_m|X_o; \theta^{t+1}), p(X_m|X_o; \theta^{t}) \right) \\ &\leq 0. 
\end{aligned}
\end{equation*}

To ensure the increase of log-likelihood in each EM iteration, we need Eq.~(\ref{eq:Q-inequality}) to hold. Using exact search methods, which can search the total parameter space, this can be guaranteed since $\theta^{t+1}=\arg \max_{\theta}\mathcal{Q}(\theta,\theta^{t})$ can be guaranteed. Then, similar to the previous conclusions by \citet{wu1983convergence,stadler2012missing, friedman1998bayesian}, MissDAG can also reach the stationary points of the overall optimization problem. However, if we use gradient-based methods (e.g., NOTEARS) to solve the optimization, this inequality can not always hold since the DAG constraint is non-convex. That is to say, we cannot guarantee to find better $\theta^{t+1}$. Even though the convergence property does not hold in this case, experimental results provided in our paper still demonstrate the effectiveness of our MissDAG.

\newpage  
\section{Supplementary experiments}
\label{app:supplementary_experiments}
\subsection{More baselines}
\label{app:more_baselines}
\begin{figure}[!ht]
\centering
\subfloat[LGM-EV.]{
  \includegraphics[width=0.95\textwidth]{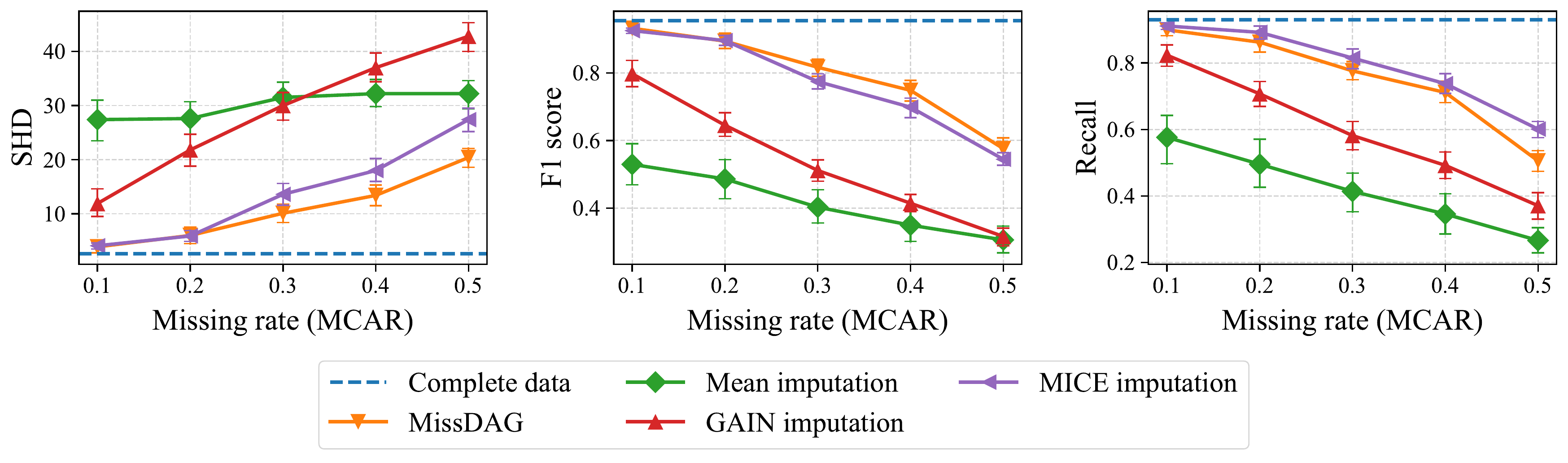}
}\\
\subfloat[LGM-NV.]{
  \includegraphics[width=0.95\textwidth]{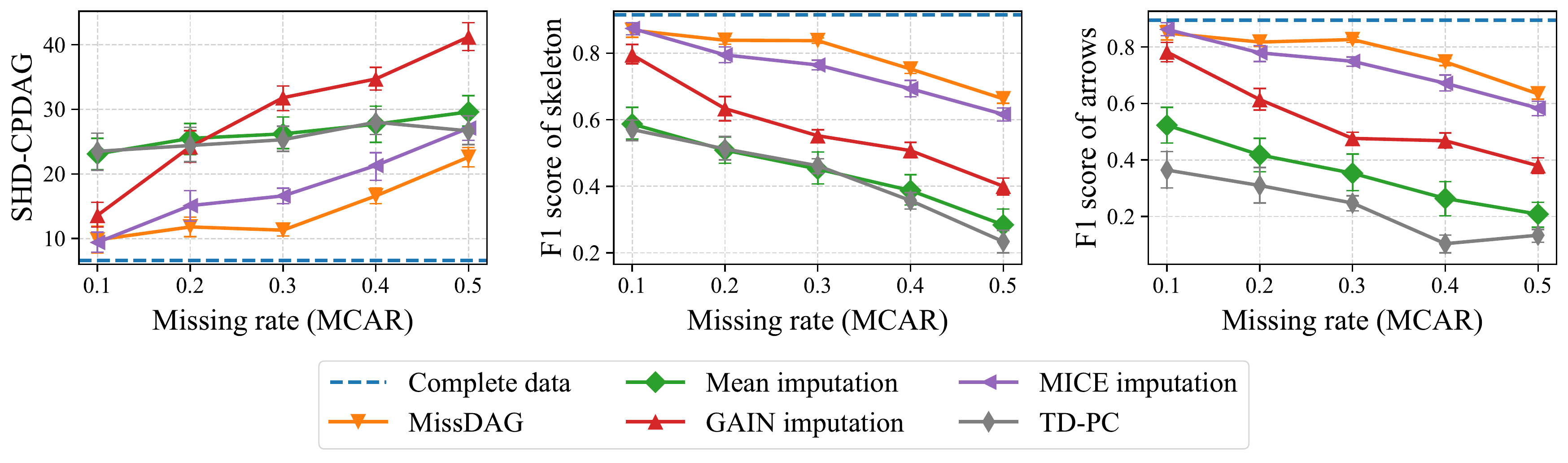}
} \\
\subfloat[LiNGAM.]{
  \includegraphics[width=0.95\textwidth]{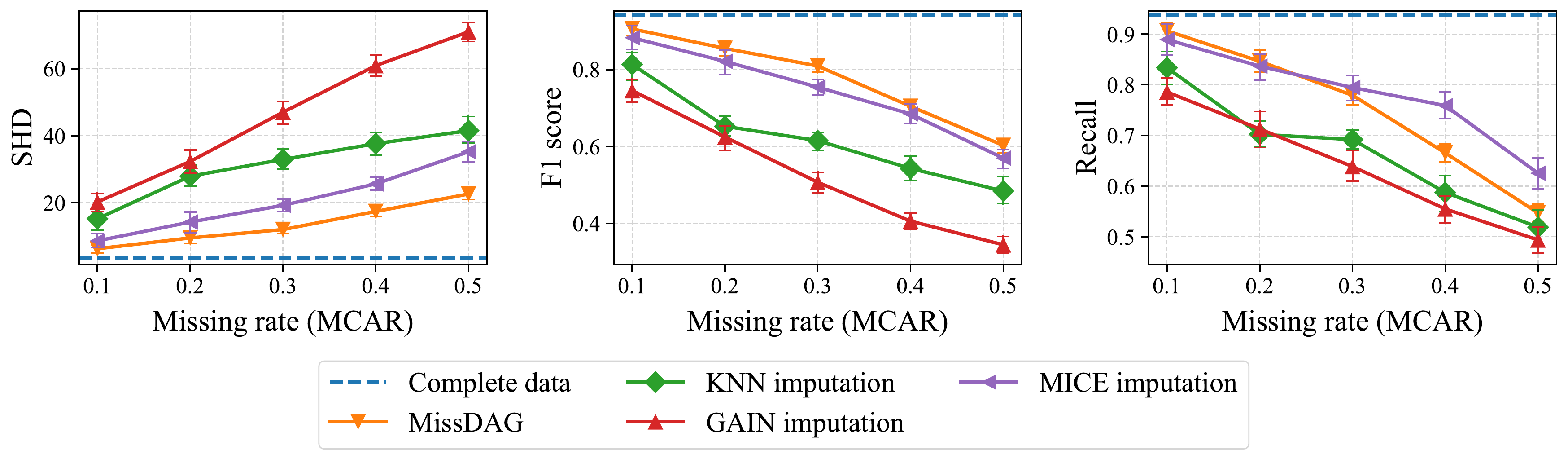}
} \\
\subfloat[NL-ANM.]{
  \includegraphics[width=0.95\textwidth]{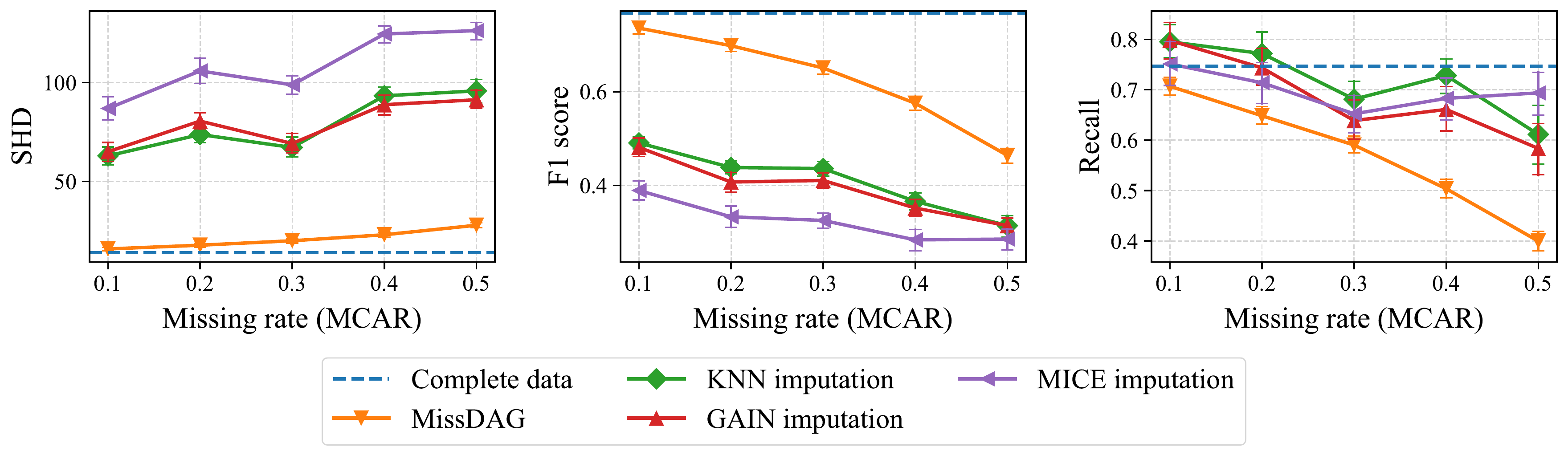}
}
\caption{Results of comparisons to more baseline methods.}
\label{fig:more_baselines}
\end{figure}

\newpage  
\subsection{Different missing types}
\label{app:different_missing_types}
\begin{figure}[!ht]
\centering
\subfloat[LGM-EV.]{
  \includegraphics[width=0.95\textwidth]{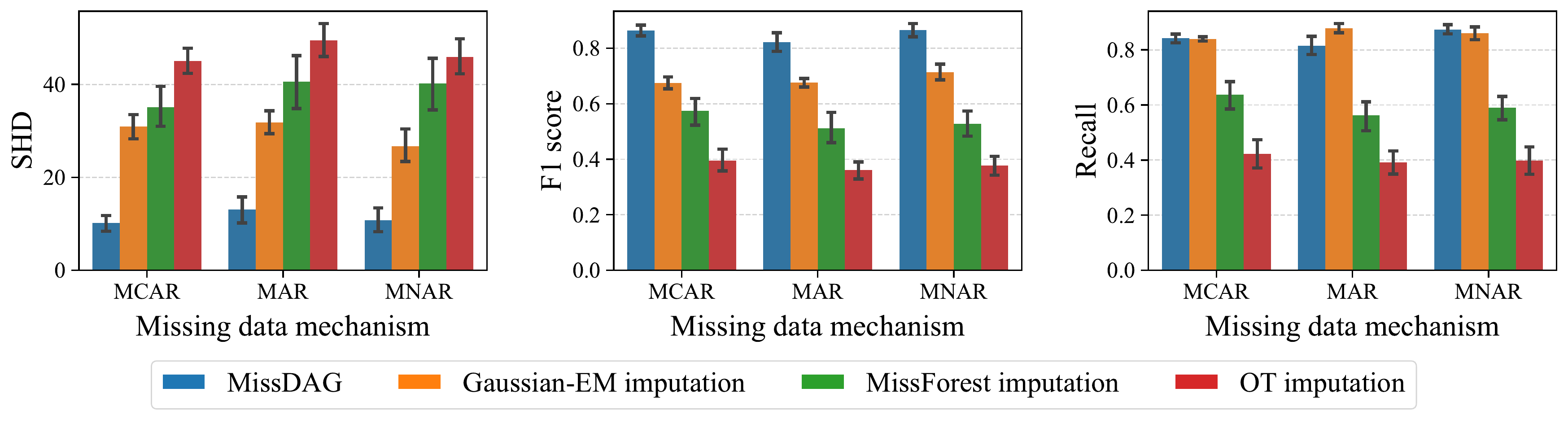}
}\\
\subfloat[LGM-NV.]{
  \includegraphics[width=0.95\textwidth]{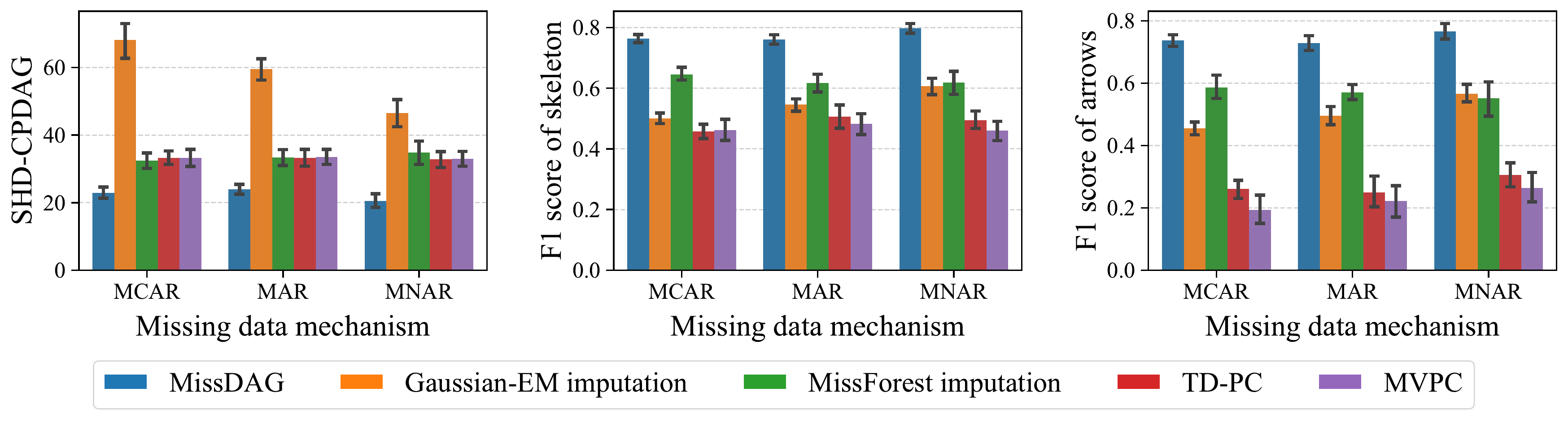}
}\\
\subfloat[LiNGAM.]{
  \includegraphics[width=0.95\textwidth]{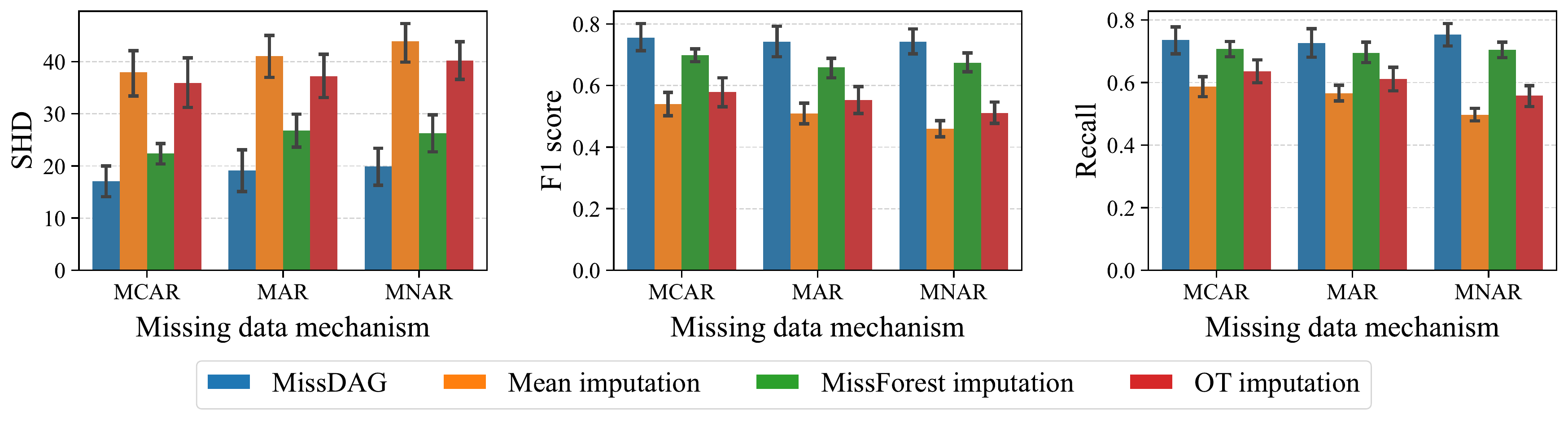}
}\\
\subfloat[NL-ANM.]{
  \includegraphics[width=0.95\textwidth]{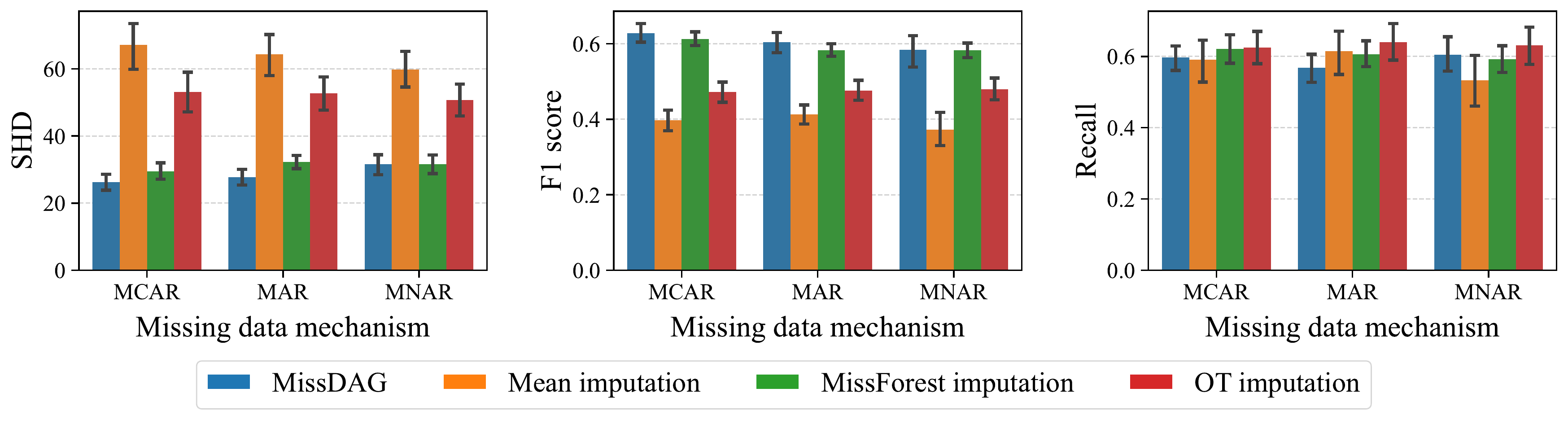}
}
\caption{Results on different missing mechanisms.}
\label{fig:different_miss_types}
\end{figure}

\newpage  
\subsection{Different numbers of samples}
\label{app:different_numbers_of_samples}
\begin{figure}[!ht]
\centering
\subfloat[LGM-EV.]{
  \includegraphics[width=0.95\textwidth]{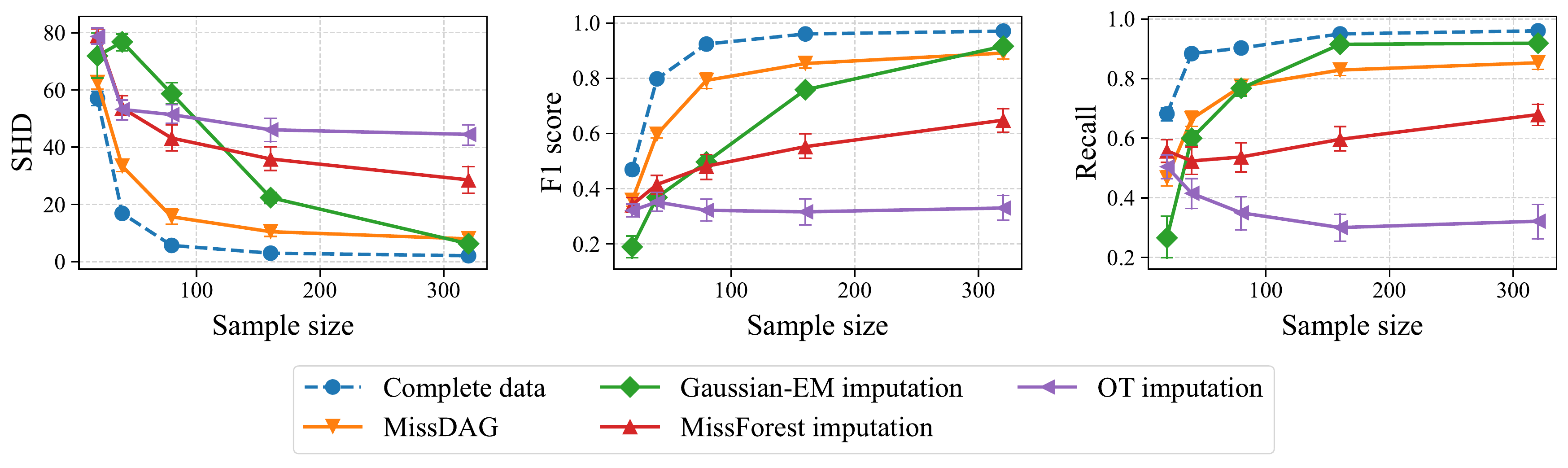}
}\\
\subfloat[LGM-NV.]{
  \includegraphics[width=0.95\textwidth]{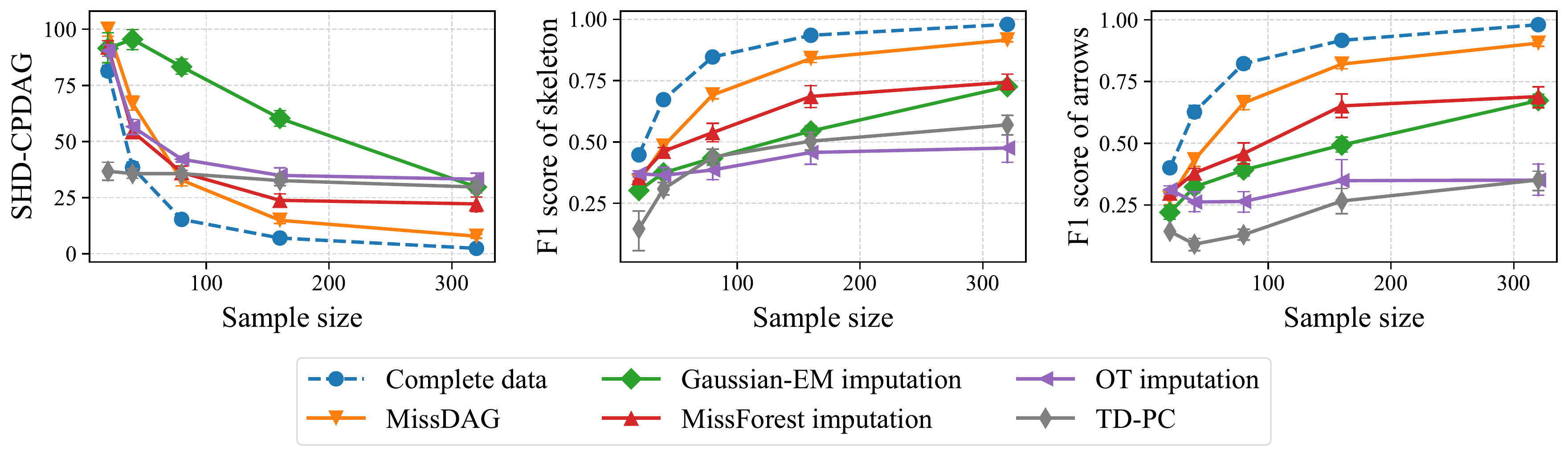}
}\\
\subfloat[LiNGAM.]{
  \includegraphics[width=0.95\textwidth]{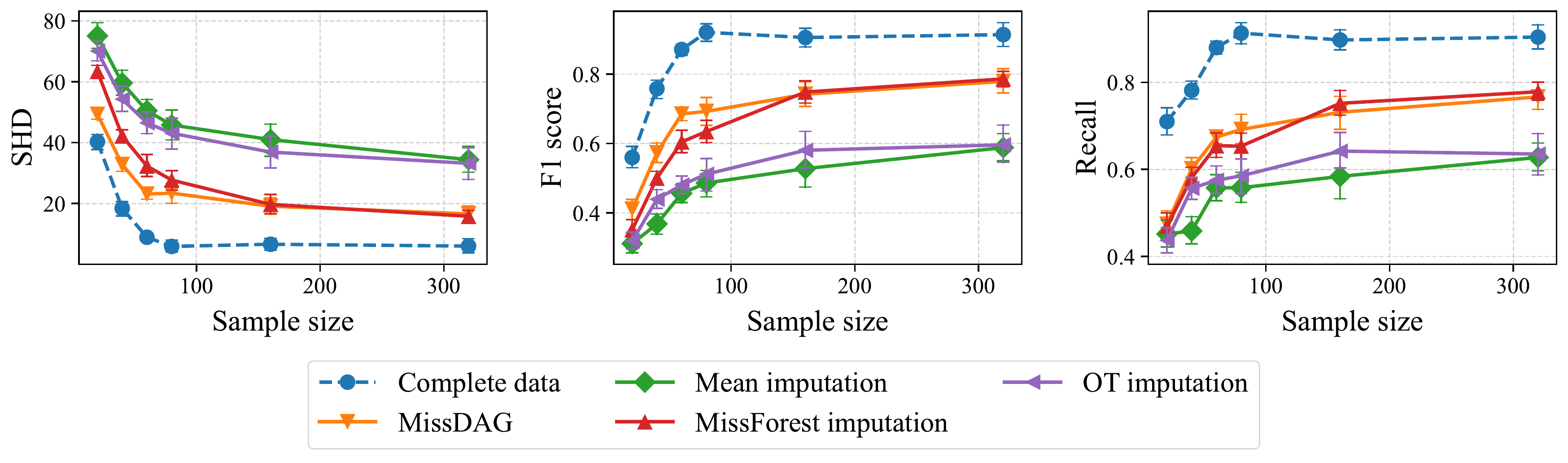}
}\\
\subfloat[NL-ANM.]{
  \includegraphics[width=0.95\textwidth]{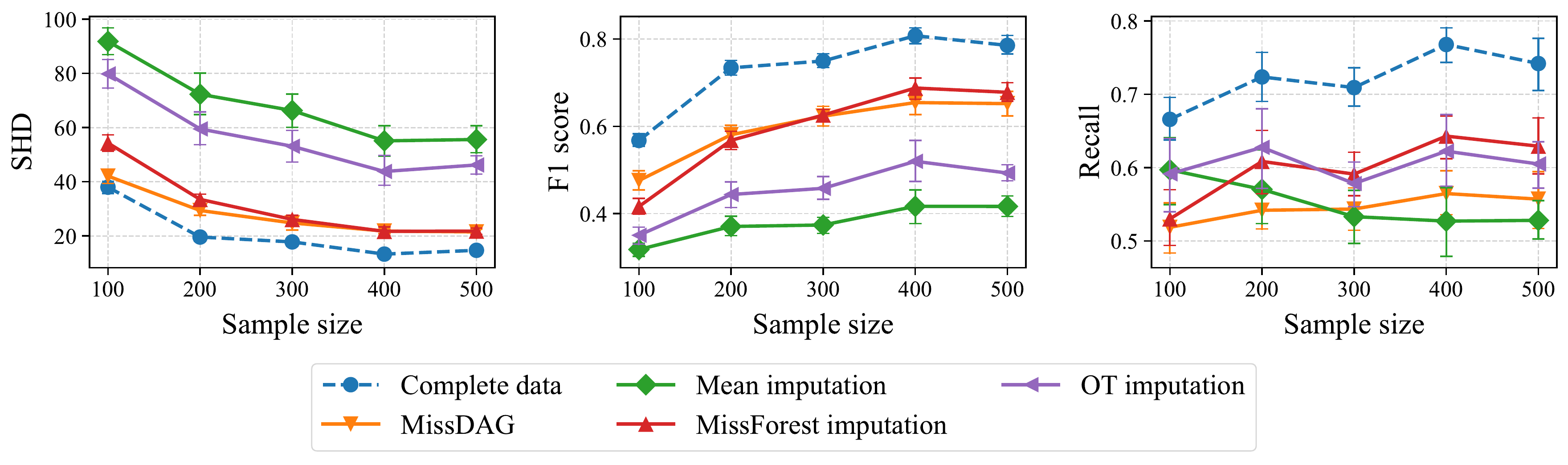}
}
\caption{Results on different number of samples.}
\label{fig:different_samples}
\end{figure}

\newpage  
\subsection{Different numbers of nodes}
\label{app:different_numbers_of_nodes}
\begin{figure}[!ht]
\centering
\subfloat[LGM-EV.]{
  \includegraphics[width=0.95\textwidth]{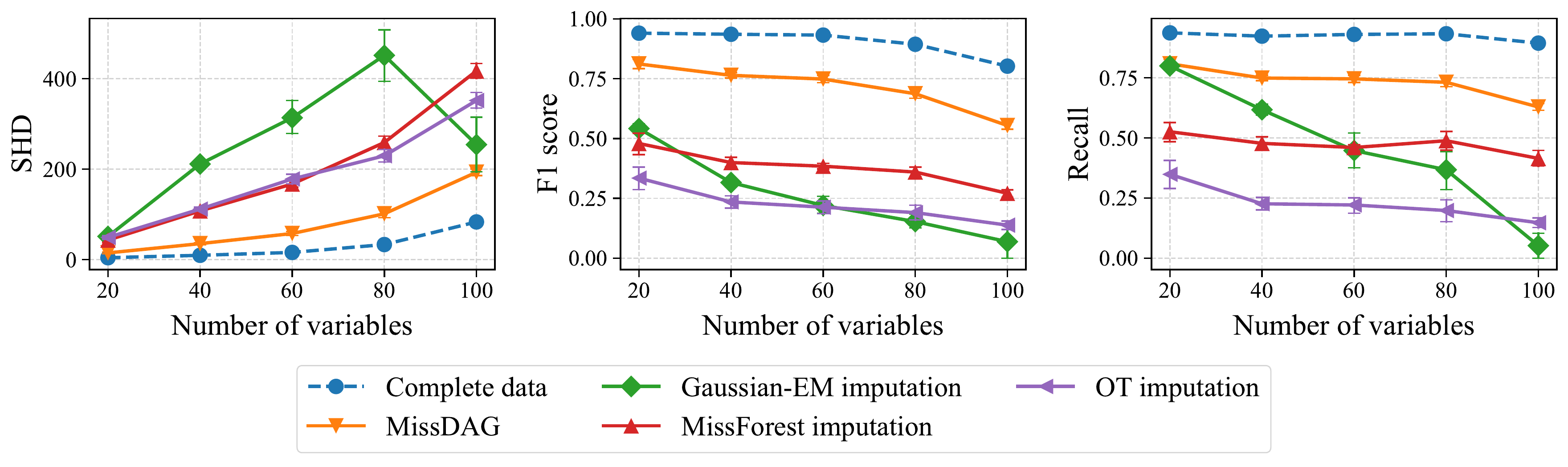}
}\\
\subfloat[LGM-NV.]{
  \includegraphics[width=0.95\textwidth]{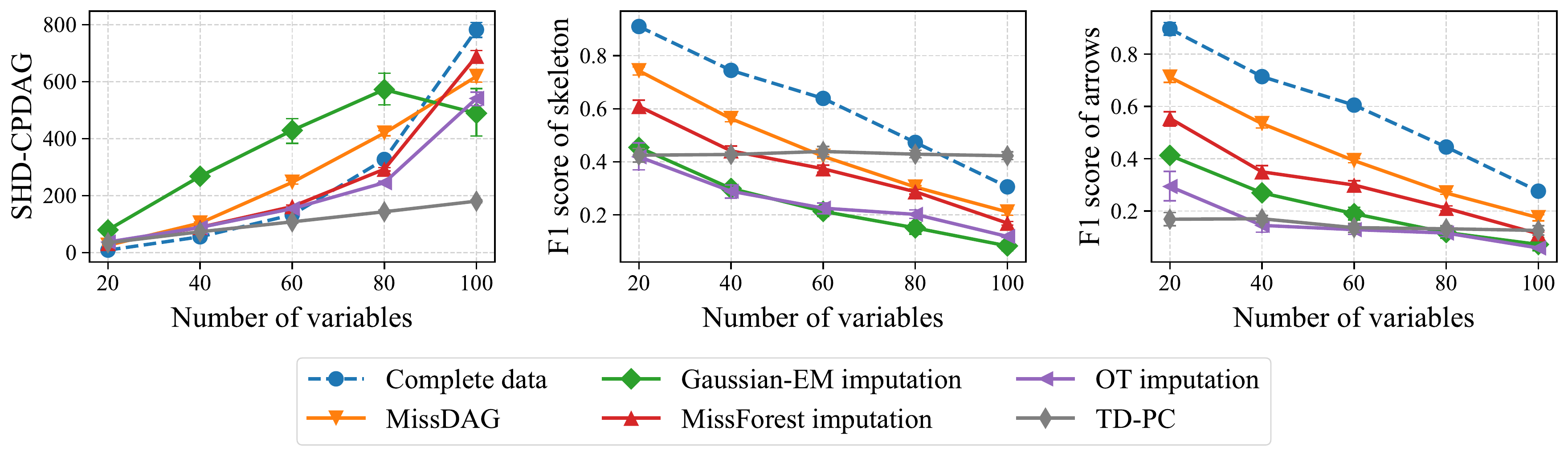}
}\\
\subfloat[LiNGAM.]{
  \includegraphics[width=0.95\textwidth]{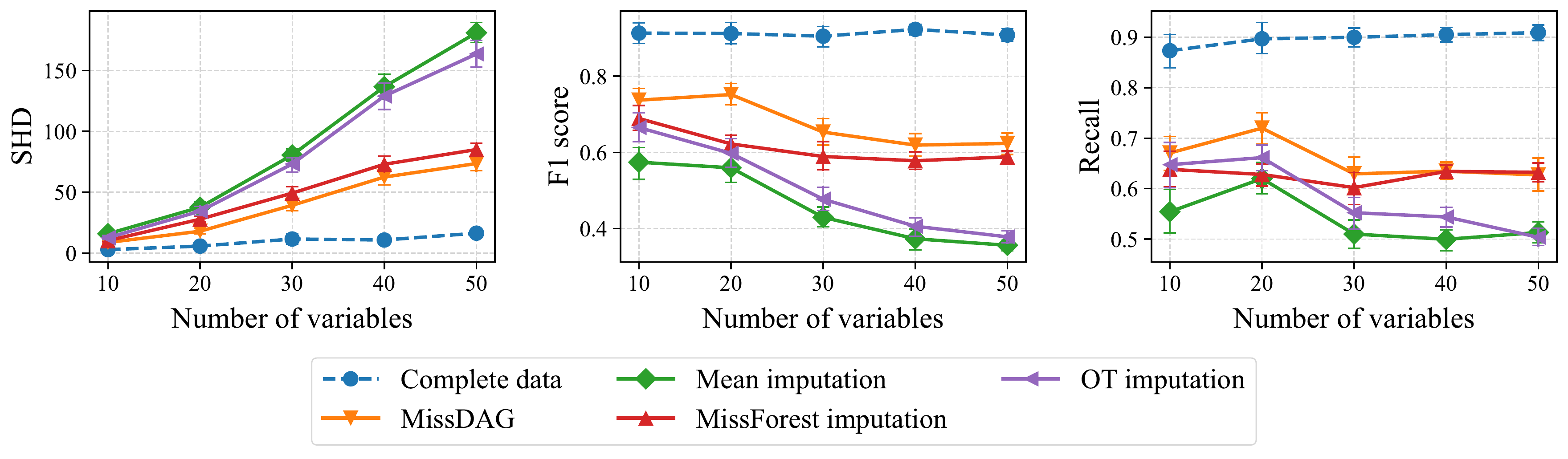}
}\\
\subfloat[NL-ANM.]{
  \includegraphics[width=0.95\textwidth]{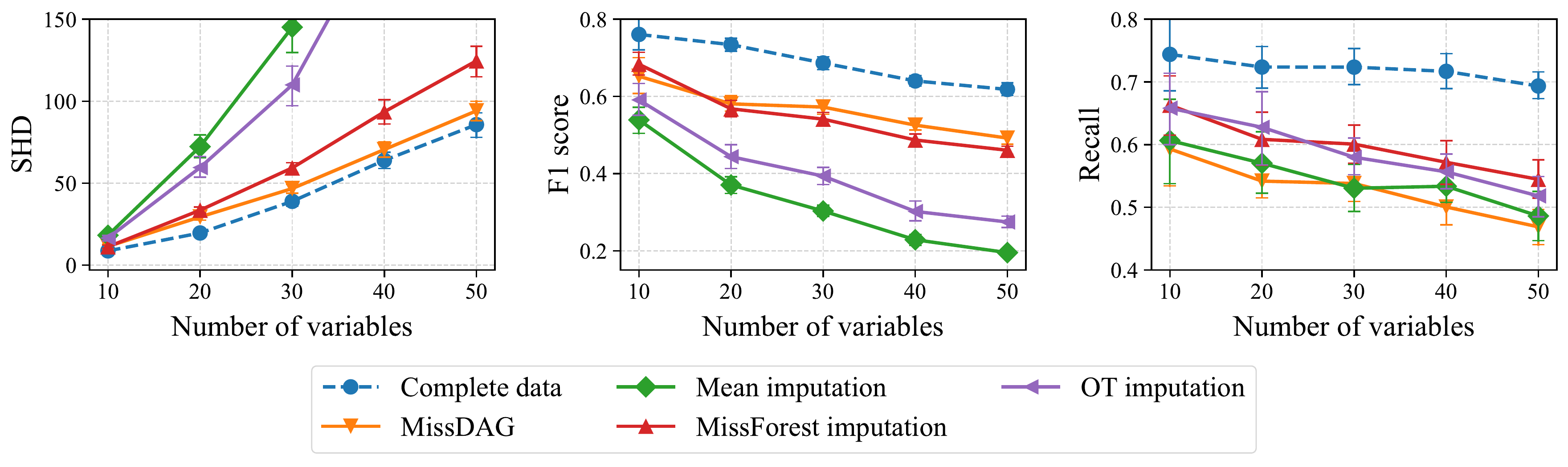}
}
\caption{Results on different numbder of nodes.}
\label{fig:different_nodes}
\end{figure}

\newpage  
\subsection{Scalability of different nodes}
\label{app:scalability_of_different_nodes}
\begin{figure}[!ht]
\centering
\subfloat[LGM-EV.]{
  \includegraphics[width=1.0\textwidth]{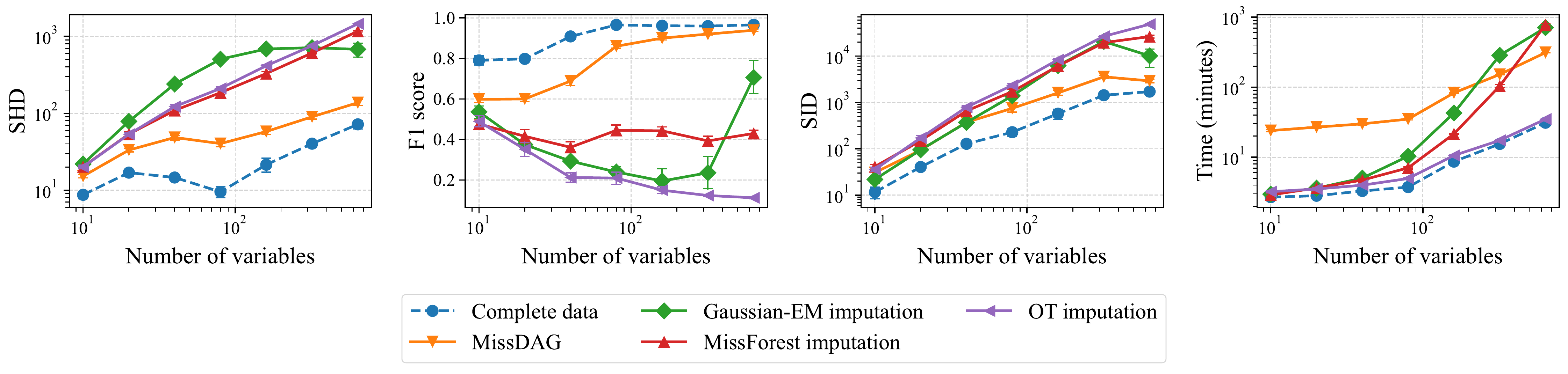}
}\\
\subfloat[LGM-NV.]{
  \includegraphics[width=1.0\textwidth]{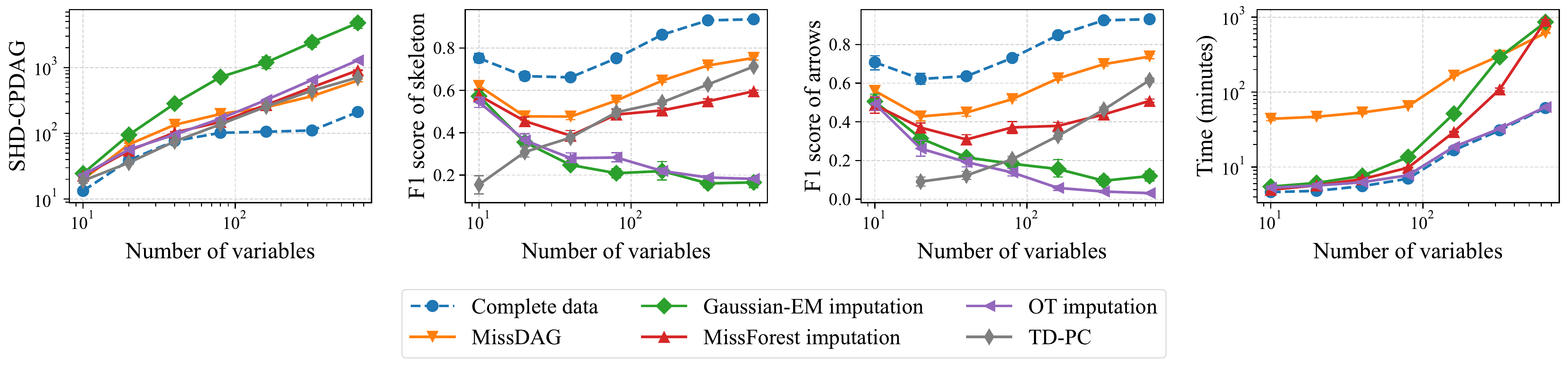}
}\\
\subfloat[LiNGAM.]{
  \includegraphics[width=0.95\textwidth]{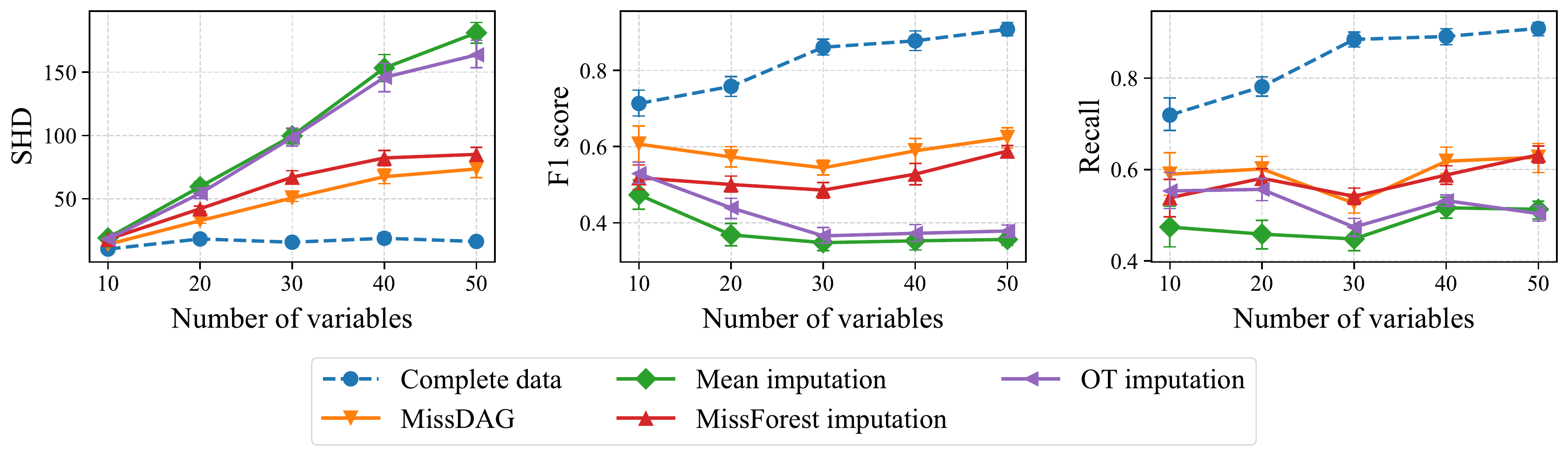}
}\\
\subfloat[NL-ANM.]{
  \includegraphics[width=0.95\textwidth]{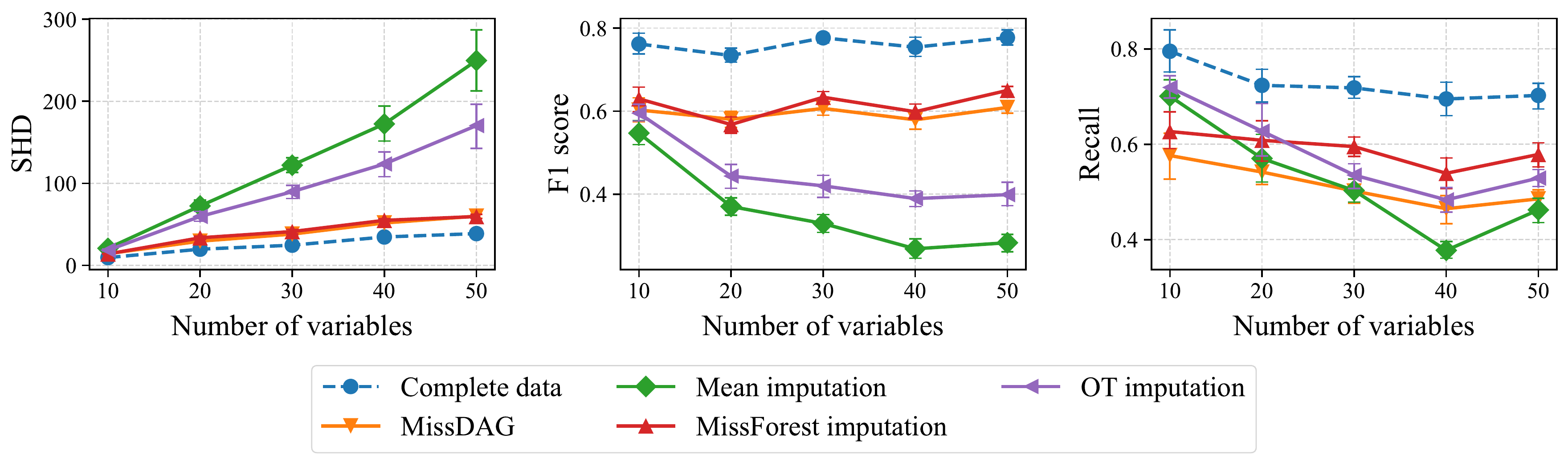}
}
\caption{Results on the scalability of different nodes ($2d$ samples for $d$ nodes). }
\label{fig:scalability}
\end{figure}


\newpage
\subsection{Different degrees}
\label{app:different_degrees}
\begin{figure}[!ht]
\centering
\subfloat[LGM-EV.]{
  \includegraphics[width=0.95\textwidth]{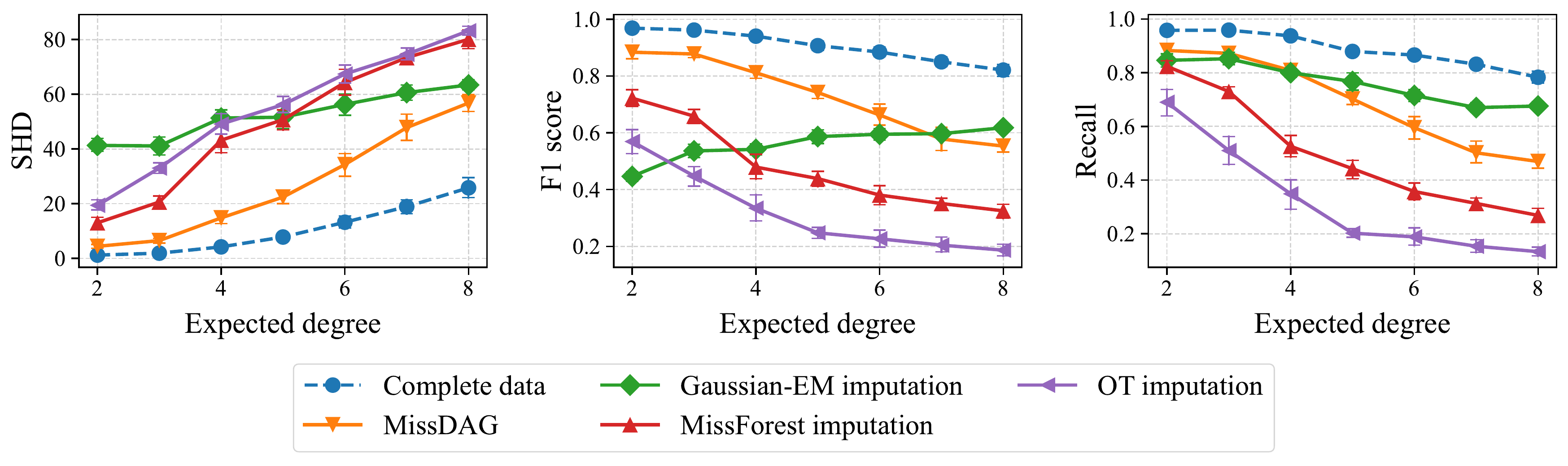}
}\\
\subfloat[LGM-NV.]{
  \includegraphics[width=0.95\textwidth]{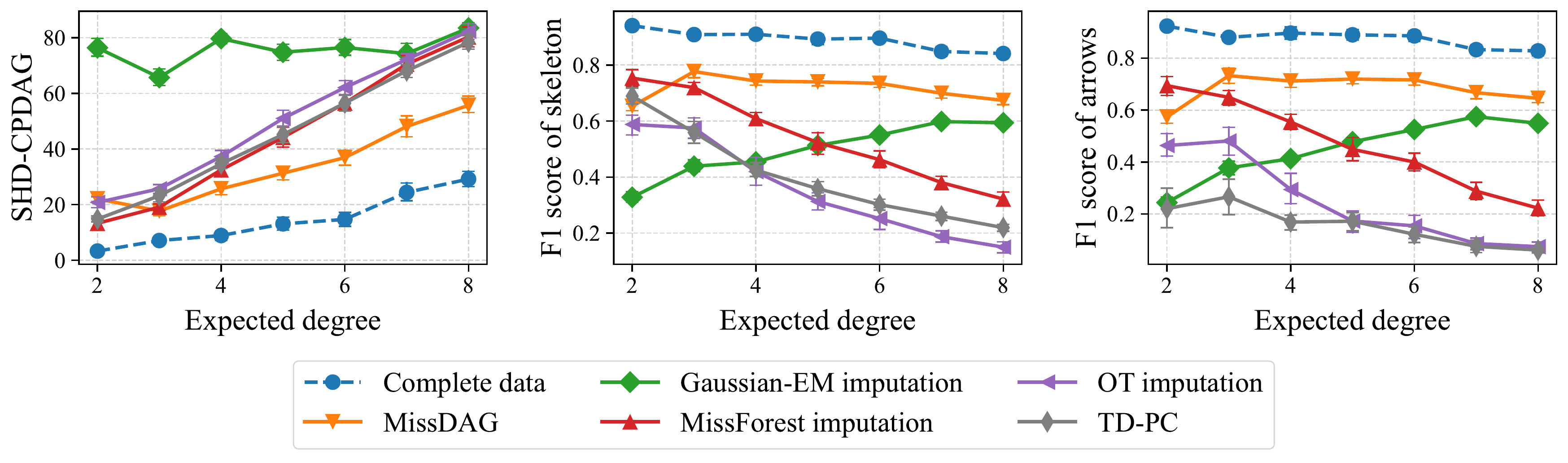}
}\\
\subfloat[LiNGAM.]{
  \includegraphics[width=0.95\textwidth]{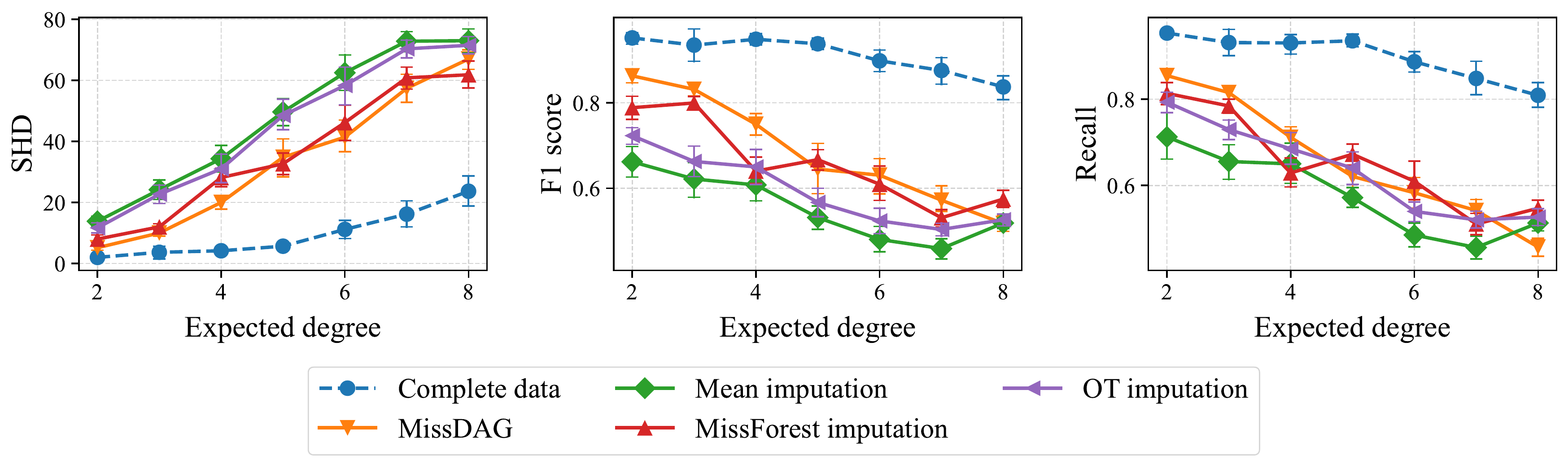}
}\\
\subfloat[NL-ANM.]{
  \includegraphics[width=0.95\textwidth]{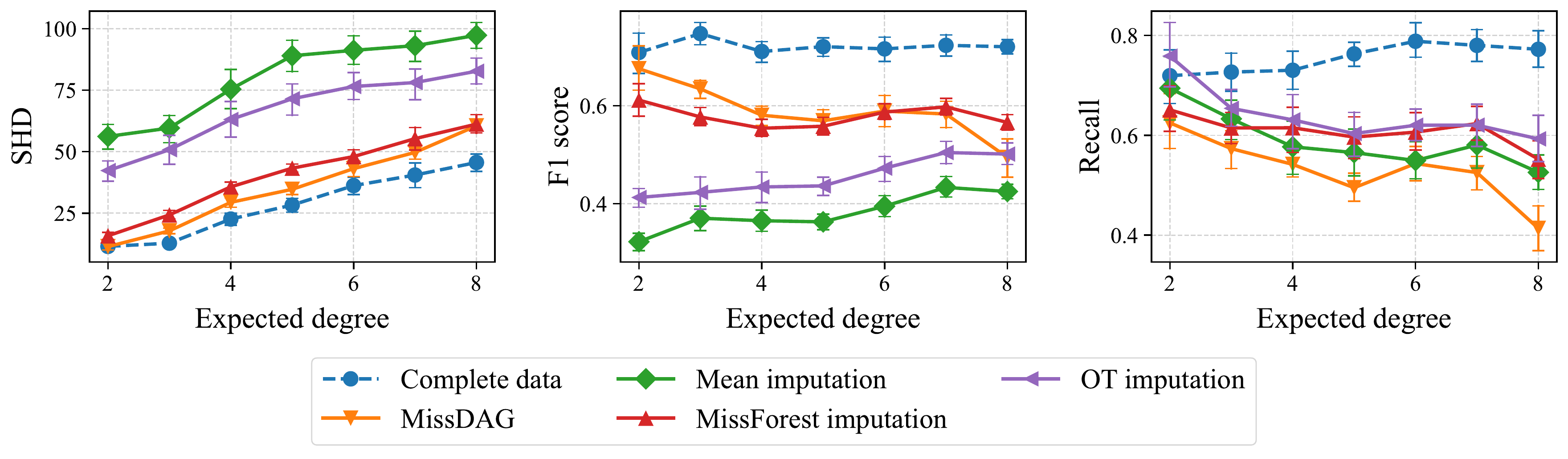}
}
\caption{Results on different degrees.}
\label{fig:different_degrees}
\end{figure}

\newpage  
\subsection{Related to LiNGAM}
\label{app:related_to_lingam}
\subsubsection{Different noise types}
\label{app:different_noise_types}
\begin{figure}[!ht]
\centering
\subfloat[Exponential.]{
\includegraphics[width=0.95\textwidth]{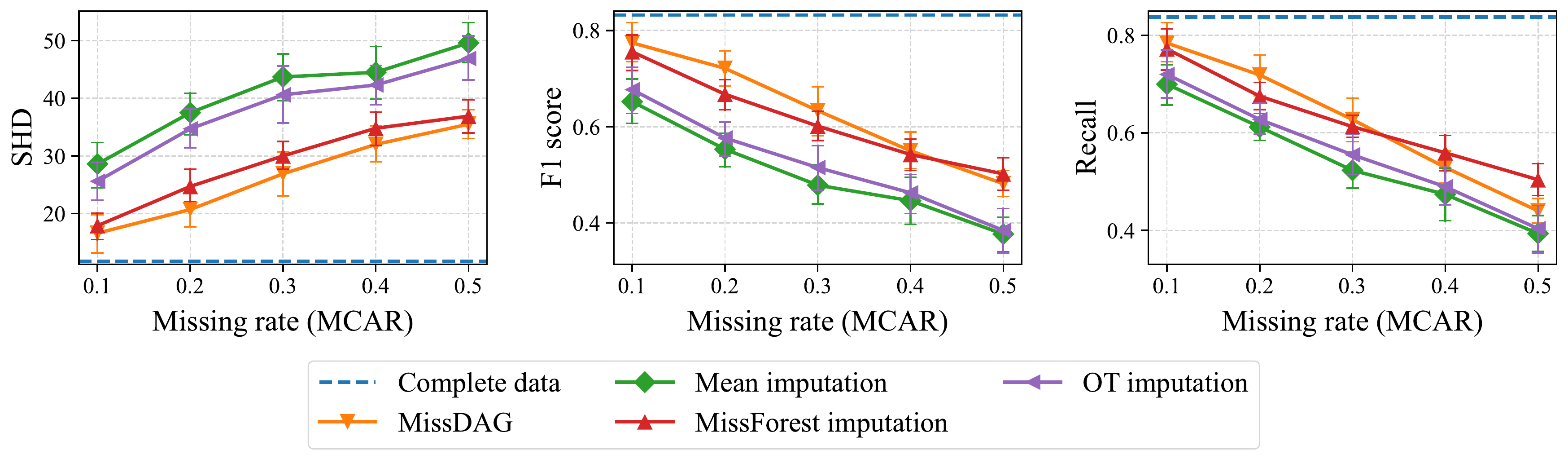}
}\\
\subfloat[Laplace.]{
\includegraphics[width=0.95\textwidth]{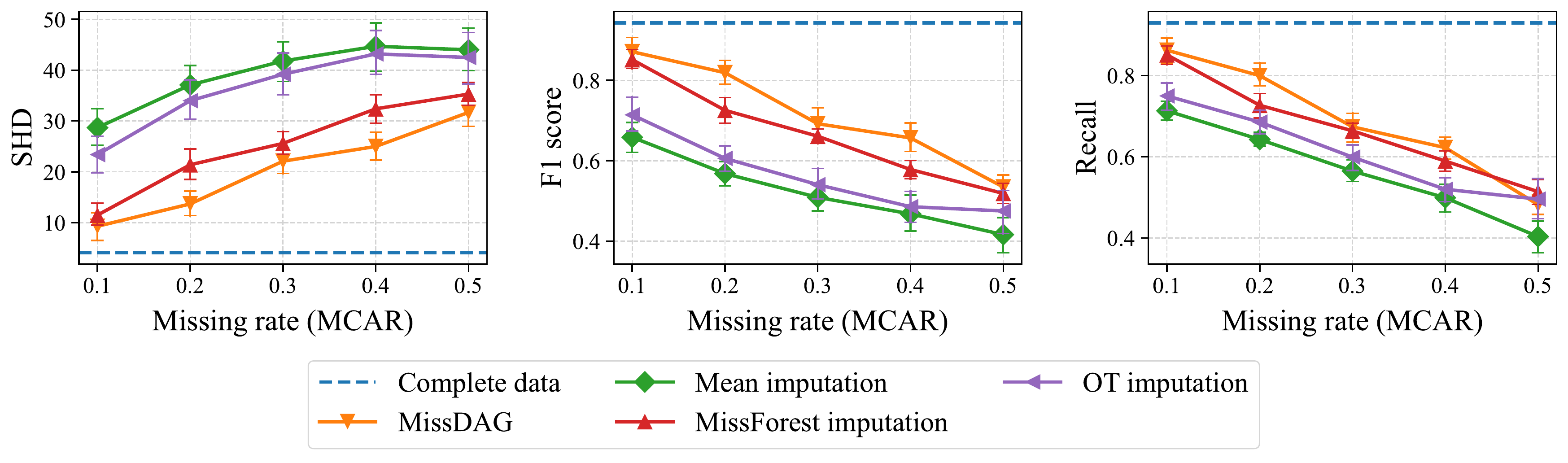}
}
\caption{Results of LiNGAM on different noise types.}
\label{fig:different_noises}
\vskip -0.1in
\end{figure}

\subsubsection{Different causal discovery methods (ICA-LiNGAM \& Direct-LiNGAM)}
\label{app:different_cd_methods}
\begin{figure}[!ht]
\centering
\subfloat[ICA-LiNGAM.]{
\includegraphics[width=0.95\textwidth]{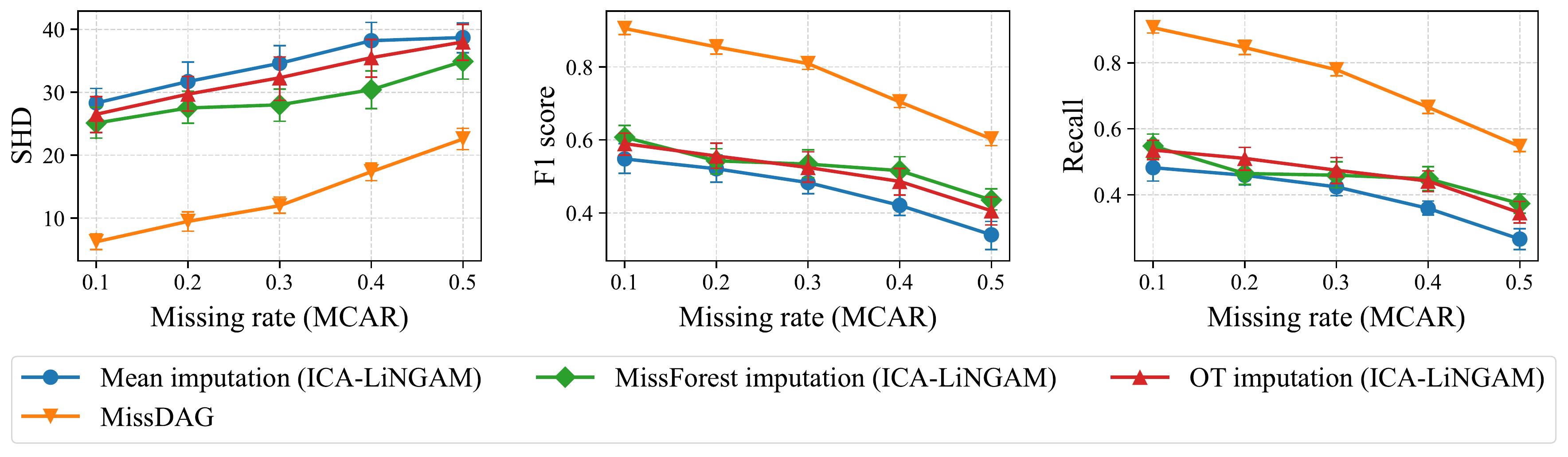}
}\\
\subfloat[Direct-LiNGAM.]{
\includegraphics[width=0.95\textwidth]{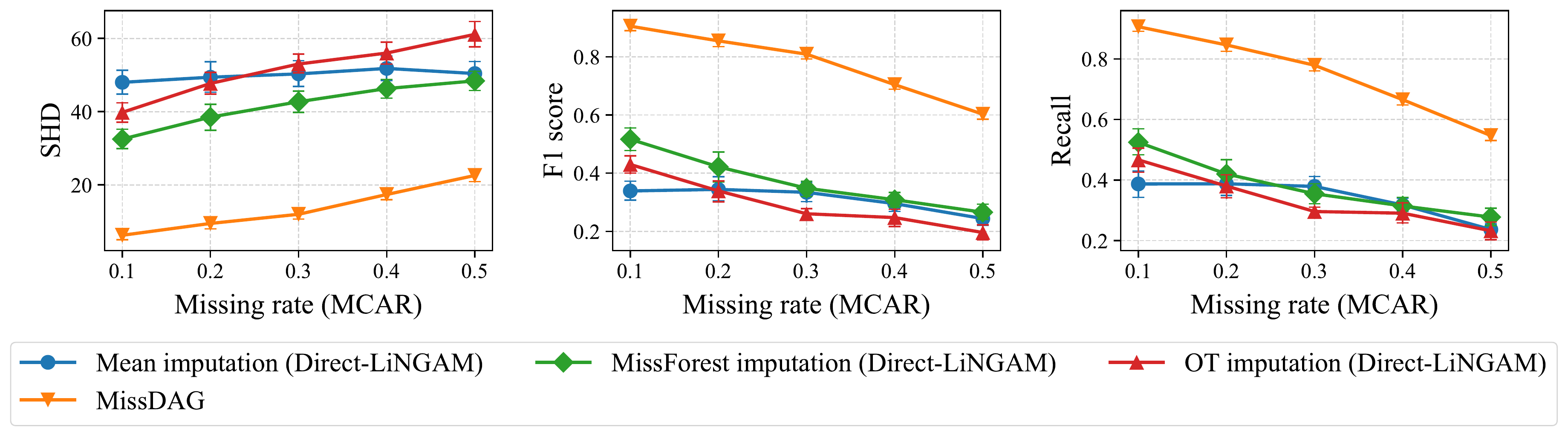}
}
\caption{Results of LiNGAM with different causal discovery baseline methods.}
\label{fig:different_cd_methods_non_gaussian}
\end{figure}

\newpage
\subsection{Different non-linear types (NL-ANM)}
\label{app:different_non_linear_types}
\begin{figure}[!ht]
\centering
\subfloat[GP-add.]{
\includegraphics[width=0.95\textwidth]{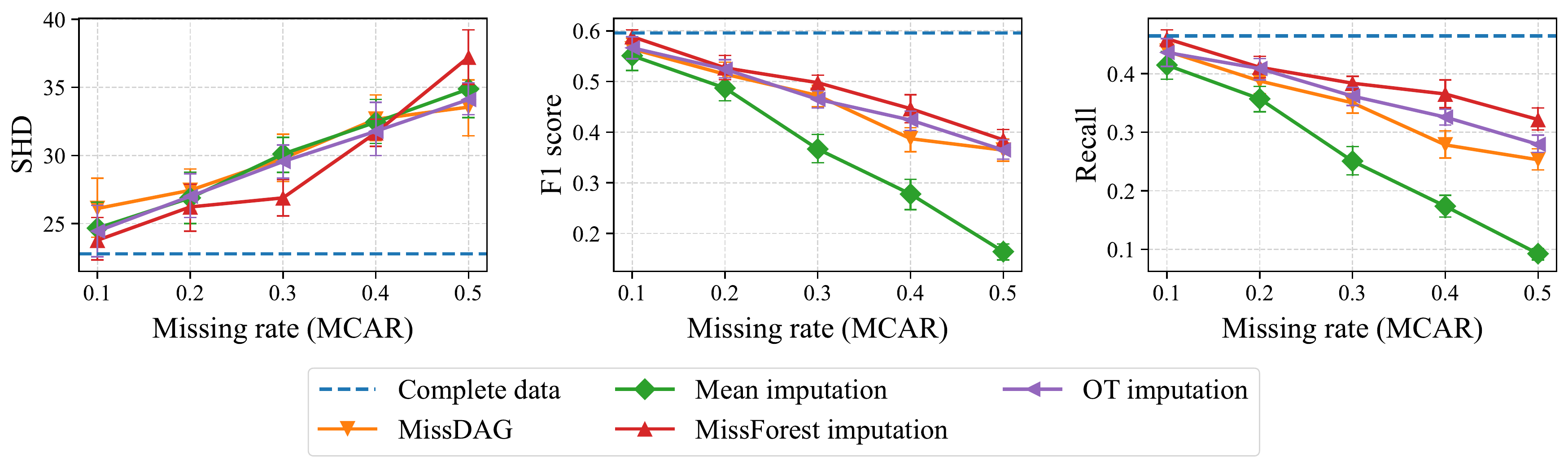}
}\\
\subfloat[MiM.]{
\includegraphics[width=0.95\textwidth]{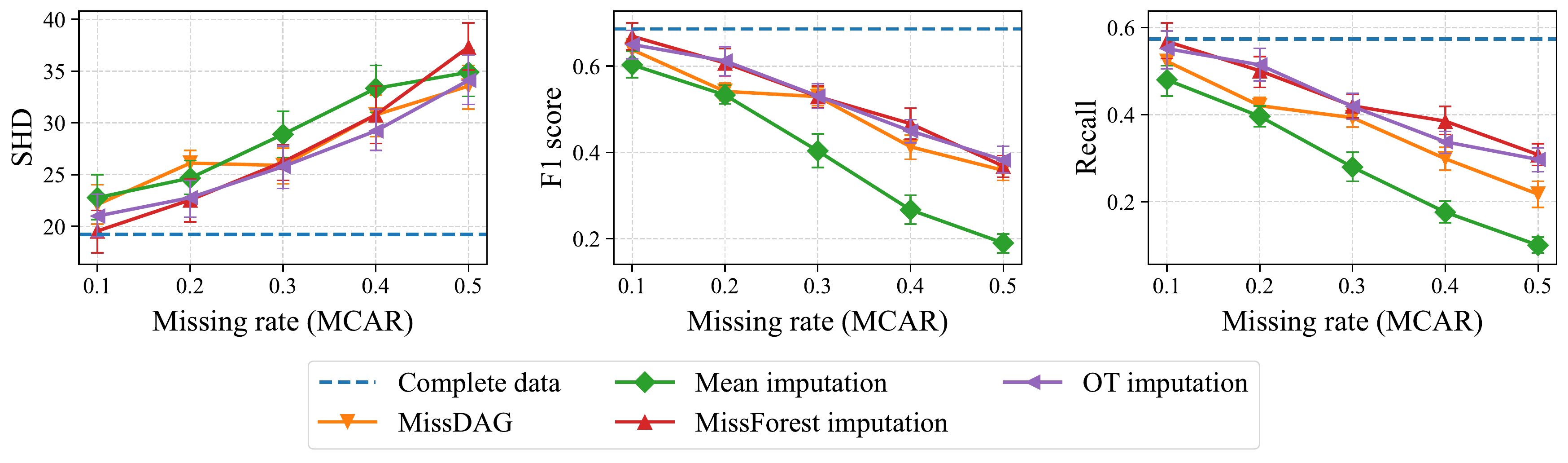}
}
\caption{Results of NL-ANM on different non-linear functions.}
\label{fig:different_non_linear_types}
\end{figure}

\section{Running time}
\label{app:running_time}
The running time of the proposed methods and the baselines are shown in Figure \ref{fig:scalability}. We observe that the proposed method has a longer running time when the number of variables is small, which may not be surprising because, different from the other one-stage imputation methods, our proposed method inherits the iterative optimization property of EM method, leading to a longer running time. Nevertheless, as soon as the number of variables gets very large (i.e., more than $640$ variables), our method runs faster than the other strong baselines (i.e., Gaussian-EM and MissForest imputations) since these methods also take much time for imputation. Moreover, the running time of the other baselines appears to increase more quickly w.r.t. the number of variables as compared to our proposed method. These observations indicate that our method appear to scale well, in addition to the improvement of structure learning performance observed in the experiments.

\end{document}